\documentclass[fleqn,10pt]{wlscirep}
\usepackage[utf8]{inputenc}
\usepackage[T1]{fontenc}
\usepackage[labelformat=simple]{subcaption}

\newcommand{\confint}[3]{#1\,(#2,\,#3)}
\usepackage{longtable}
\usepackage{listings}
\usepackage{array}
\usepackage{tabularx}
\usepackage{multirow}
\usepackage{xcolor}
\definecolor{DeepGreen}{rgb}{0.0, 0.5, 0.0}
\usepackage[normalem]{ulem}

\title{MedReadCtrl: Personalizing medical text generation with readability-controlled instruction learning}

\author[1,2+]{Hieu Tran, MSc}
\author[1,2+]{Zonghai Yao, MSc}
\author[1,3]{Won Seok Jang, MSc, RN}
\author[1,3]{Sharmin Sultana, BSc}
\author[4]{Allen Chang, MD}
\author[5]{Yuan Zhang, Ph.D., RN}
\author[1,2,3,4*]{Hong Yu, PhD}

\affil[1]{Center for Healthcare Organization and Implementation Research, VA Bedford Health Care, MA, USA}
\affil[2]{Manning College of Information and Computer Sciences, UMass Amherst, MA, USA}
\affil[3]{Miner School of Computer and Information Sciences, UMass Lowell, MA, USA}
\affil[4]{Department of Medicine, University of Massachusetts Medical School, Worcester, MA, USA}
\affil[5]{School of Nursing, Zuckerberg College of Health Sciences, UMass Lowell, MA, USA}

\affil[*]{Corresponding author: Hong Yu (Hong\_Yu@uml.edu)}

\affil[+]{these authors contributed equally to this work}

\begin{abstract}
Generative AI has demonstrated strong potential in healthcare, from clinical decision support to patient-facing chatbots that improve outcomes. 
A critical challenge for deployment is effective human–AI communication, where content must be both personalized and understandable. 
We introduce MedReadCtrl, a readability-controlled instruction tuning framework that enables LLMs to adjust output complexity without compromising meaning.
Evaluations of nine datasets and three tasks across medical and general domains show that MedReadCtrl achieves significantly lower readability instruction-following errors than GPT-4 (e.g., 1.39 vs. 1.59 on ReadMe, p<0.001) and delivers substantial gains on unseen clinical tasks (e.g., +14.7 ROUGE-L, +6.18 SARI on MTSamples). 
Experts consistently preferred MedReadCtrl (71.7\% vs. 23.3\%), especially at low literacy levels.
These gains reflect MedReadCtrl’s ability to restructure clinical content into accessible, readability-aligned language while preserving medical intent, offering a scalable solution to support patient education and expand equitable access to AI-enabled care.

\end{abstract}
\begin{document}

\maketitle

\section{Introduction}
\label{Intro}

The rapid evolution of generative artificial intelligence (AI) is introducing a new era of highly personalized and seamless human-machine communication~\cite{golan2024rise}. 
In healthcare, this transformation has the potential to fundamentally redefine clinical care by tailoring interactions and information delivery to meet the unique needs, preferences, and comprehension levels of individual patients~\cite{rajpurkar2022ai,silcox2024potential,yao2025survey}. 
At its core, a key challenge for AI in healthcare is to maximize its communicative effectiveness with humans—empowering patients, improving engagement, and ultimately advancing patient-centered care~\cite{sauerbrei2023impact,robinson2024reviewing}.

Achieving this vision, however, requires overcoming a critical and persistent barrier: the vast disparities in domain-specific literacy between users, ranging from patients with limited health literacy to professionals in specialized fields such as law, finance, or medicine.
Policies such as Open Medical Records and initiatives such as Blue Button~\cite{delbanco2010open,BlueButton2024} exemplify ongoing efforts to democratize access to medical information and promote transparency, engagement, and shared decision-making~\cite{nutbeam2023artificial}. 
However, the benefits of these policies are often limited by the patient's ability to comprehend complex medical records and health-related information~\cite{khasawneh2022effect}. 
Without ensuring that the information is understandable, accessibility alone cannot translate into meaningful empowerment.

The emergence of large language models (LLMs) introduces unprecedented opportunities to bridge this communication gap~\cite{meng2024application}.
LLMs excel at complex medical reasoning and content generation, and their growing integration into healthcare systems offers the promise of scalable, personalized patient education and support~\cite{achiam2023gpt,kung2023performance,yang2023performance}.
However, for these systems to realize their full potential, it is not enough for AI to generate clinically accurate information; the content must also be tailored to individual patients’ literacy levels, backgrounds, and preferences.

In this work, we focus specifically on patient education, a critical component of patient–provider communication that centers on delivering health-related information in a manner patients can understand and act upon.
While patient communication encompasses a broader range of interactions, including dialogue, emotional support, and shared decision-making, patient education refers more narrowly to the informational aspect—providing explanations about diagnoses, treatments, and health behaviors.
Effective patient education is essential for improving comprehension, adherence, and health outcomes, especially when tailored to individuals’ health literacy levels and backgrounds.

In this context, readability-controlled text generation emerges as a key enabler of personalization, which not only helps providers educate patients more efficiently, but also provides important support for communication and mutual assistance among patients, especially within groups with similar diseases or support groups.
This truly enables information to flow among patients, promoting shared decision-making and self-management.


However, existing efforts in controllable text generation are largely limited to binary simplification, complication, or style-transfer tasks~\cite{yao2024readme,guo2021automated,luo2022readability,yao2021improving,guo2023appls}. 
These approaches fall short of meeting the nuanced and diverse personalization needs of real-world healthcare users, largely due to the limitations of available training data and model adaptability.
To address these challenges, the field is shifting towards more dynamic, user-adaptive systems capable of fine-grained output control~\cite{liang2024controllable}. 
This adaptability is especially crucial in healthcare, where demographic and socioeconomic factors significantly influence patients' health literacy and comprehension~\cite{vo2023association}. 
Static, one-size-fits-all approaches to readability are insufficient.
Instead, there is a growing demand for AI systems that can dynamically adjust communication complexity to fit each patient's unique context, thereby enhancing engagement, shared decision-making, and self-management~\cite{shi2024general}.

In response, we propose Medical Readability-Controlled Instruction Learning (MedReadCtrl), a novel framework designed to endow LLMs with fine-grained readability control capabilities specifically tailored for healthcare contexts.
Our approach integrates explicit instruction tuning based on targeted readability levels, enabling LLMs to transform input text into outputs that align with patients’ comprehension abilities. 
We systematically evaluate our system, LlaMA3-MedReadCtrl, across various tasks—text simplification, paraphrase generation, and semantic entailment generation—each crafted to assess the model’s ability to adjust output complexity while preserving semantic fidelity.
Our extensive automatic and human evaluations across nine datasets in both medical and general domains demonstrate that MedReadCtrl significantly improves LLMs’ performance in readability-controlled generation, outperforming existing models and laying a strong foundation for more personalized, accessible healthcare communication.

\noindent Our contributions are summarized as follows:
\begin{itemize}
  \item We propose MedReadCtrl, a novel medical readability-controlled instruction learning framework that enhances LLMs' ability to generate content with controlled readability, making it more suitable for patient education and engagement.
  \item To the best of our knowledge, LlaMA3-MedReadCtrl is the first system specifically designed to address readability-controlled text generation in the medical domain, enabling fine-grained control over text complexity across different readability levels.
  \item Extensive automatic and human evaluations validate LlaMA3-MedReadCtrl's effectiveness in both the medical and general NLP domains, demonstrating its ability to improve LLM's performance across three key tasks (text simplification, paraphrase generation, and semantic entailment generation) on nine datasets. This highlights its potential for enhancing patient education by ensuring readability-controlled medical text generation.
\end{itemize}

\begin{figure*}[!htbp]
\vspace{-10mm}
    \centering
    \begin{subfigure}{\textwidth}
        \centering
        \includegraphics[width=\textwidth]{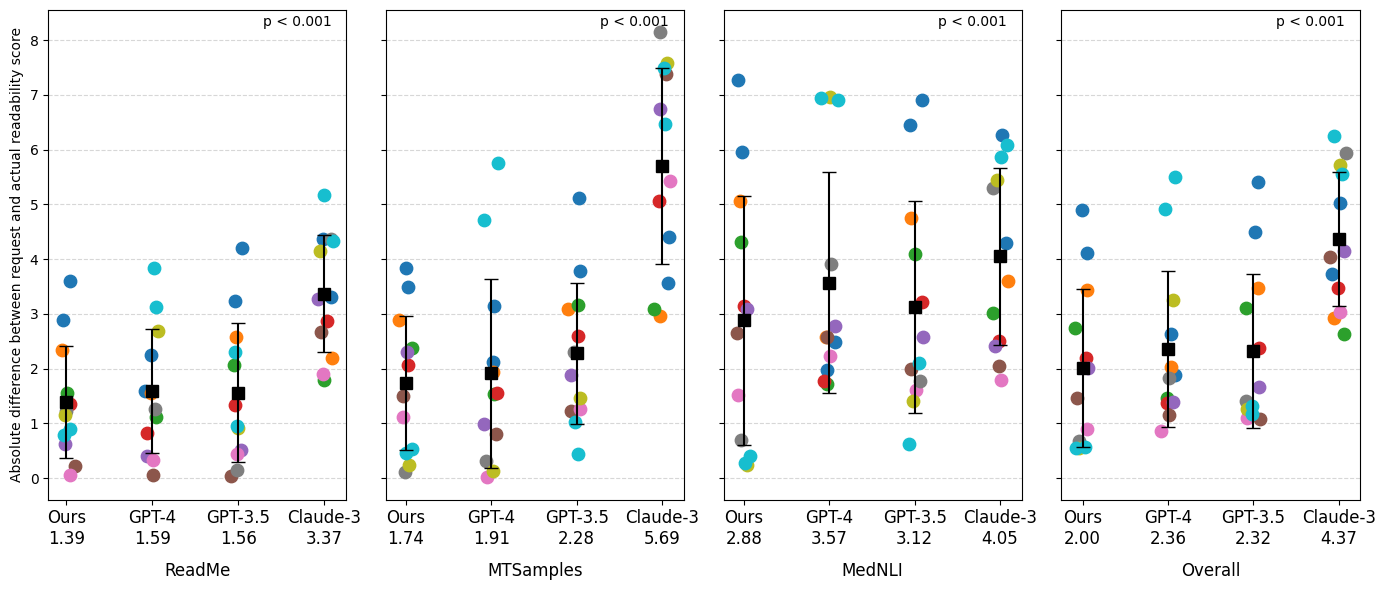}
        \vspace{-3mm}
        \caption{Comparative analysis of readability control capabilities between different models.
        The results are measured using the absolute mean difference across four common readability metrics. 
        Each point represents the difference between the request grade and the actual-generated readability (12 points for Grades 1-12).
        The median and interquartile range for each variable are plotted. 
        The reported P-values correspond to results from nonparametric Mann-Whitney tests.
        Across three medical-related datasets (ReadMe, MTSamples, MedNLI), Ours consistently achieves lower readability control error (Delta) compared to GPT-4: ReadMe (1.39 vs. 1.59, p < 0.001), MTSamples (1.74 vs. 1.91, p < 0.001), MedNLI (2.88 vs. 3.57, p < 0.001), and Overall (2.00 vs. 2.36, p < 0.001).
        Similar trends are observed when comparing our system to GPT-3.5 and Claude 3 across all datasets, with consistently lower Delta scores (all p < 0.001).}
        \label{subfig:confidential}
    \end{subfigure}
    
    \vspace{1em} 

    \begin{subfigure}{0.48\textwidth}
        \centering
        \includegraphics[width=0.9\textwidth]{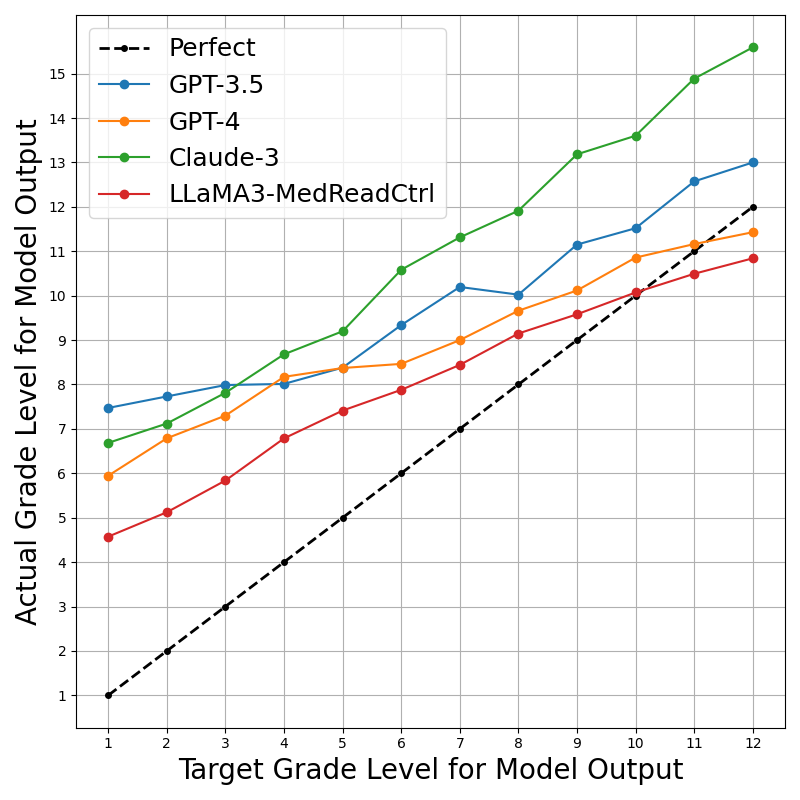}
        \caption{MedReadCtrl instruction following on the general domain}
        \label{subfig:tasktype}
    \end{subfigure}
    \hfill
    \begin{subfigure}{0.48\textwidth}
        \centering
        \includegraphics[width=0.9\textwidth]{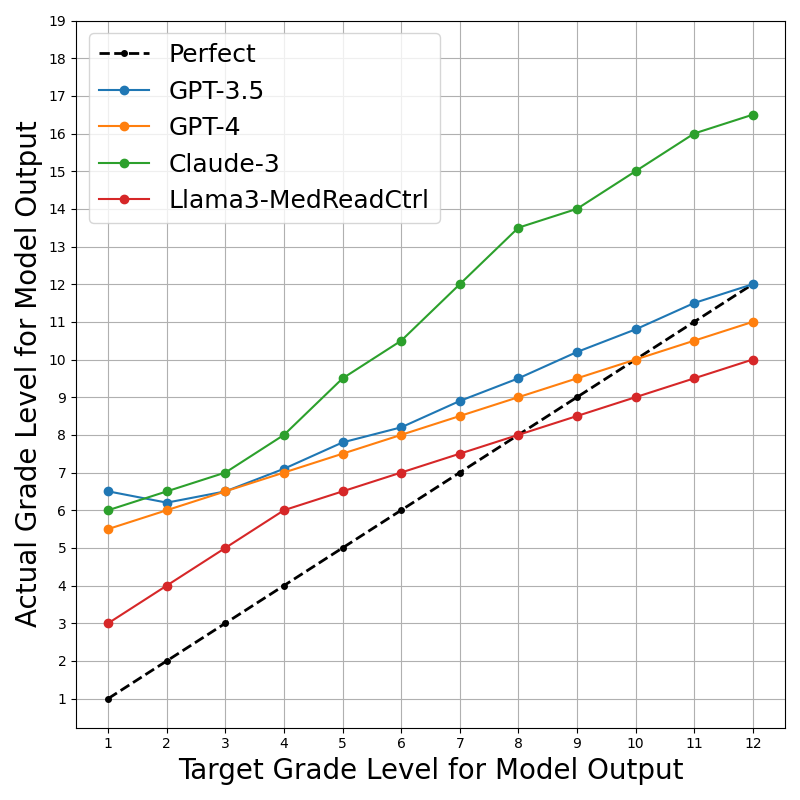}
        \caption{MedReadCtrl instruction following on the medical domain}
        \label{subfig:overview}
    \end{subfigure}
    \vspace{2mm}
    \caption{Comparison of readability control precision between our LLAMA3-MedReadCtrl model and GPT-4. (a) LLAMA3-MedReadCtrl demonstrates significantly lower errors in readability control across medical datasets.
    (b) presents the ability of different LLMs to follow readability-control instructions in general-domain texts. The y-axis represents the model's actual readability output, while the x-axis denotes the target readability level. GPT-4, GPT-3.5, and Claude-3 exhibit an overall upward trend, indicating partial adherence to readability-control instructions, but their curves deviate significantly from the ideal (Perfect) curve, with errors increasing at higher readability levels. In contrast, LLAMA3-8B (without MedReadCtrl training) shows a nearly flat trend, suggesting minimal readability-control capability. LLAMA3-MedReadCtrl closely follows the ideal curve, demonstrating superior readability-control instruction-following ability and precise readability adjustment.
    (c) evaluates different LLMs on readability-control instruction-following ability in medical texts. GPT and Claude models exhibit patterns similar to their general-domain performance, showing moderate readability-control capability but substantial deviations from the ideal curve, especially at higher target readability levels where errors increase. 
    LLAMA3-MedReadCtrl significantly outperforms other models, closely aligning with the ideal curve, highlighting its strong and stable readability-control ability in the medical domain.
    }
    \label{fig:ReadCtrl_instruction_following_ability}
\end{figure*}

\renewcommand{\arraystretch}{0.82} 

\begin{table*}[!ht]
\centering
\scalebox{1}{
\begin{tabular}{lcccc}
\toprule
Models & ROUGE-1 & ROUGE-L & BLEU & SARI \\
\midrule
\multicolumn{5}{c}{\textbf{Medical Domain Tasks}} \\
\hline
\multicolumn{5}{l}{\textbf{ReadMe (seen) -- Medical Text Simplification}} \\[4pt]
Claude-3 & \confint{34.77}{34.00}{35.54} & \confint{30.66}{29.90}{31.42} & \confint{07.25}{06.50}{08.00} & \confint{51.39}{50.50}{52.28} \\
GPT-3.5  & \confint{35.67}{35.10}{36.44} & \confint{30.69}{29.90}{31.48} & \confint{08.32}{07.55}{09.09} & \confint{50.27}{49.40}{51.14} \\
GPT-4 & \confint{38.48}{37.70}{39.26} & \confint{27.46}{26.70}{28.22} & \confint{06.36}{05.60}{07.12} & \confint{51.39}{50.50}{52.28} \\
Llama3-MedReadCtrl & \textbf{\confint{39.93}{39.16}{40.70}} & \textbf{\confint{30.79}{30.00}{31.58}} & \textbf{\confint{10.24}{09.45}{11.03}} & \textbf{\confint{53.52}{52.65}{54.39}} \\
\midrule
\multicolumn{5}{l}{\textbf{MedNLI (seen) -- Semantic Entailment Generation}} \\[4pt]
Claude-3 & \confint{26.23}{25.46}{26.99} & \confint{25.28}{24.52}{26.04} & \confint{02.76}{02.00}{03.52} & \confint{53.92}{53.08}{54.76} \\
GPT-3.5 & \confint{29.91}{29.14}{30.68} & \confint{29.08}{28.32}{29.84} & \confint{04.59}{03.83}{05.35} & \confint{56.57}{55.70}{57.44} \\
GPT-4 & \confint{24.68}{23.92}{25.44} & \confint{23.90}{23.14}{24.66} & \confint{01.78}{01.02}{02.54} & \confint{56.12}{55.26}{56.98} \\
Llama3-MedReadCtrl & \textbf{\confint{40.27}{39.50}{41.04}} & \textbf{\confint{39.49}{38.72}{40.26}} & \textbf{\confint{10.35}{09.59}{11.11}} & \textbf{\confint{63.97}{63.10}{64.84}} \\
\midrule
\multicolumn{5}{l}{\textbf{MTSamples (unseen) -- Medical Text Simplification}} \\[4pt]
Claude-3 & \confint{37.93}{37.16}{38.70} & \confint{34.75}{33.98}{35.52} & \confint{10.77}{10.00}{11.54} & \confint{18.97}{18.20}{19.74} \\
GPT-3.5 & \confint{46.85}{46.08}{47.62} & \confint{42.28}{41.51}{43.05} & \confint{16.11}{15.34}{16.88} & \confint{19.22}{18.45}{19.99} \\
GPT-4 & \confint{39.40}{38.63}{40.17} & \confint{36.44}{35.67}{37.21} & \confint{11.63}{10.86}{12.40} & \confint{17.03}{16.26}{17.80} \\
Llama3-MedReadCtrl & \textbf{\confint{53.99}{53.22}{54.76}} & \textbf{\confint{51.14}{50.37}{51.91}} & \textbf{\confint{29.73}{28.96}{30.50}} & \textbf{\confint{23.21}{22.44}{23.98}} \\

\midrule
\multicolumn{5}{c}{\textbf{General Domain Tasks}} \\
\hline
\multicolumn{5}{l}{\textbf{ASSET (seen) -- Text Simplification}} \\[4pt]
Claude-3 & \confint{53.84}{53.30}{54.38} & \confint{47.29}{45.96}{48.62} & \confint{18.74}{17.64}{19.84} & \confint{40.70}{39.80}{41.60} \\
GPT-3.5 & \confint{59.76}{58.50}{61.02} & \confint{52.42}{51.26}{53.58} & \confint{27.39}{26.81}{27.97} & \confint{41.01}{40.10}{41.92} \\
GPT-4 & \confint{54.87}{53.97}{55.77} & \confint{48.26}{47.54}{48.98} & \confint{20.61}{19.40}{21.82} & \confint{39.73}{38.74}{40.72} \\
Llama3-MedReadCtrl & \confint{\textbf{66.09}}{\textbf{65.38}}{\textbf{66.80}} & \confint{\textbf{59.16}}{\textbf{57.80}}{\textbf{60.52}} & \confint{\textbf{38.56}}{\textbf{37.27}}{\textbf{39.85}} & \confint{\textbf{45.60}}{\textbf{44.44}}{\textbf{46.76}} \\
\midrule
\multicolumn{5}{l}{\textbf{SNLI (seen) -- Semantic Entailment Generation}} \\[4pt]
Claude-3 & \confint{33.96}{32.97}{34.95} & \confint{31.81}{31.04}{32.58} & \confint{4.46}{3.13}{5.79} & \confint{48.33}{47.51}{49.15} \\
GPT-3.5 & \confint{38.08}{36.84}{39.32} & \confint{34.81}{33.56}{36.06} & \confint{8.75}{8.21}{9.29} & \confint{51.02}{49.88}{52.16} \\
GPT-4 & \confint{39.32}{37.90}{40.74} & \confint{36.18}{35.56}{36.80} & \confint{10.50}{9.33}{11.67} & \confint{\textbf{52.12}}{\textbf{50.87}}{\textbf{53.37}} \\
Llama3-MedReadCtrl & \confint{\textbf{41.77}}{\textbf{40.55}}{\textbf{42.99}} & \confint{\textbf{37.53}}{\textbf{36.37}}{\textbf{38.69}} & \confint{\textbf{11.22}}{\textbf{10.05}}{\textbf{12.39}} & \confint{49.93}{48.74}{51.12} \\
\midrule
\multicolumn{5}{l}{\textbf{PAWS (seen) -- Paraphrase Generation}} \\[4pt]
Claude-3 & \confint{57.21}{56.11}{58.31} & \confint{50.83}{49.82}{51.84} & \confint{23.93}{22.89}{24.97} & \confint{38.35}{37.38}{39.32} \\
GPT-3.5 & \confint{62.93}{61.51}{64.35} & \confint{52.87}{51.41}{54.33} & \confint{38.73}{37.87}{39.59} & \confint{37.98}{37.41}{38.55} \\
GPT-4 & \confint{75.37}{74.44}{76.30} & \confint{70.64}{69.38}{71.90} & \confint{31.22}{30.37}{32.07} & \confint{34.35}{33.04}{35.66} \\
LLaMA3-MedReadCtrl & \confint{\textbf{93.09}}{\textbf{92.25}}{\textbf{93.93}} & \confint{\textbf{84.34}}{\textbf{82.99}}{\textbf{85.69}} & \confint{\textbf{66.49}}{\textbf{65.55}}{\textbf{67.43}} & \confint{\textbf{60.53}}{\textbf{59.05}}{\textbf{62.01}} \\
\midrule
\multicolumn{5}{l}{\textbf{WikiSmall (unseen) -- Text Simplification}} \\[4pt]
Claude-3        & \confint{48.74}{48.00}{49.48} & \confint{40.43}{39.70}{41.16} & \confint{16.06}{15.30}{16.82} & \confint{33.00}{32.30}{33.70} \\
GPT-3.5         & \confint{50.18}{49.42}{50.94} & \confint{44.69}{43.94}{45.44} & \confint{19.40}{18.64}{20.16} & \confint{33.98}{33.24}{34.72} \\
GPT-4           & \confint{46.66}{45.92}{47.40} & \confint{40.23}{39.50}{40.96} & \confint{16.66}{15.90}{17.42} & \confint{31.47}{30.72}{32.22} \\
Llama3-MedReadCtrl & \confint{\textbf{61.35}}{\textbf{60.60}}{\textbf{62.10}} & \confint{\textbf{55.86}}{\textbf{55.12}}{\textbf{56.60}} & \confint{\textbf{37.28}}{\textbf{36.54}}{\textbf{38.02}} & \confint{\textbf{40.55}}{\textbf{39.80}}{\textbf{41.30}} \\
\midrule
\multicolumn{5}{l}{\textbf{MultiNLI (unseen) -- Semantic Entailment Generation}} \\[4pt]
Claude-3        & \confint{30.13}{29.38}{30.88} & \confint{27.23}{26.50}{27.96} & \confint{03.03}{02.30}{03.76} & \confint{44.43}{43.68}{45.18} \\
GPT-3.5         & \confint{31.81}{31.06}{32.56} & \confint{26.19}{25.44}{26.94} & \confint{03.78}{03.03}{04.53} & \confint{44.06}{43.32}{44.80} \\
GPT-4           & \confint{33.48}{32.73}{34.23} & \confint{29.91}{29.16}{30.66} & \confint{05.62}{04.87}{06.37} & \confint{\textbf{46.42}}{\textbf{45.68}}{\textbf{47.16}} \\
Llama3-MedReadCtrl & \confint{\textbf{40.27}}{\textbf{39.52}}{\textbf{41.02}} & \confint{\textbf{33.49}}{\textbf{32.75}}{\textbf{34.23}} & \confint{\textbf{11.37}}{\textbf{10.63}}{\textbf{12.11}} & \confint{43.83}{43.08}{44.58} \\
\midrule
\multicolumn{5}{l}{\textbf{MRPC (unseen) -- Paraphrase Generation}} \\[4pt]
Claude-3        & \confint{41.54}{40.80}{42.28} & \confint{37.12}{36.38}{37.86} & \confint{16.79}{16.05}{17.53} & \confint{36.78}{36.04}{37.52} \\
GPT-3.5         & \confint{44.93}{44.18}{45.68} & \confint{38.23}{37.48}{38.98} & \confint{20.59}{19.84}{21.34} & \confint{37.34}{36.60}{38.08} \\
GPT-4           & \confint{\textbf{66.82}}{\textbf{66.06}}{\textbf{67.58}} & \confint{\textbf{61.39}}{\textbf{60.64}}{\textbf{62.14}} & \confint{15.30}{14.56}{16.04} & \confint{34.85}{34.10}{35.60} \\
LLaMA3-MedReadCtrl  & \confint{64.68}{63.93}{65.43} & \confint{58.47}{57.72}{59.22} & \confint{\textbf{37.98}}{\textbf{37.23}}{\textbf{38.73}} & \confint{\textbf{44.43}}{\textbf{43.68}}{\textbf{45.18}} \\
\bottomrule
\end{tabular}
}
\caption{Main results for MedReadCtrl on general and medical domain tasks. All values are shown as mean (lower bound, upper bound), representing 95\% confidence intervals.}
\label{tab:merged_results}
\end{table*}

\section{Results}
\label{results}

\subsection{Automatic evaluation results}

Figure~\ref{fig:ReadCtrl_instruction_following_ability} presents a comprehensive evaluation of the readability control precision of our LLAMA3-MedReadCtrl model compared with GPT-4, GPT-3.5, Claude-3, and a LLAMA3 baseline across both general and medical domains.
In Figure~\ref{fig:ReadCtrl_instruction_following_ability}a, our LLAMA3-MedReadCtrl model demonstrates significantly improved precision in readability control relative to GPT-4 across multiple medical datasets. Specifically, the model achieves substantially lower average absolute readability errors (computed across four standard readability metrics: Gunning Fog, Flesch-Kincaid, Automated Readability Index, Coleman-Liau):
\begin{itemize}
\setlength\itemsep{-0.5em}
    \item \textbf{ReadMe dataset:} LLAMA3-MedReadCtrl (1.39) significantly outperformed GPT-4 (1.59, \textit{p}$<$0.001).
    \item \textbf{MTSamples dataset:} LLAMA3-MedReadCtrl (1.74) significantly outperformed GPT-4 (1.91, \textit{p}$<$0.001).
    \item \textbf{MedNLI dataset:} LLAMA3-MedReadCtrl (2.88) significantly outperformed GPT-4 (3.57, \textit{p}$<$0.001).
\end{itemize}
\setlength\itemsep{0em}
These results demonstrate that our system provides superior readability calibration tailored for medical applications.
In general-domain readability tasks (Figure~\ref{fig:ReadCtrl_instruction_following_ability}b), LLAMA3-MedReadCtrl consistently follows the ideal readability level (Perfect curve). In contrast, GPT-4, GPT-3.5, and Claude-3 show increasing deviations as the requested readability level rises, indicating their limited capability to precisely adhere to readability-level instructions. Notably, the baseline LLAMA3 model without MedReadCtrl training fails to demonstrate meaningful readability-level adherence, highlighting the essential role of our readability control training.
Similar patterns appear within medical-domain instruction-following evaluations (Figure~\ref{fig:ReadCtrl_instruction_following_ability}c). LLAMA3-MedReadCtrl closely aligns with targeted readability levels, whereas GPT-4, GPT-3.5, and Claude-3 deviate increasingly at higher readability targets. This disparity underscores the suitability of LLAMA3-MedReadCtrl for precise readability control within specialized medical contexts, significantly enhancing its utility for clinical communication tasks where accurate readability adjustments are crucial.

Table~\ref{tab:merged_results} summarizes the automatic evaluation results of LLAMA3-MedReadCtrl compared to Claude-3, GPT-3.5, and GPT-4 on medical and general domain text generation tasks, using standard NLP metrics (ROUGE, BLEU, and SARI).
Across medical domain datasets, LLAMA3-MedReadCtrl consistently achieved the highest performance, clearly outperforming all other evaluated models, both in seen and unseen scenarios—where seen refers to datasets (e.g., MedNLI) used during the readability-controlled instruction tuning phase, and unseen refers to datasets (e.g., MTSamples) that were not included in our training pipeline and are used to assess the model’s generalization ability. Note that since base models like LLaMA3 are pretrained on large-scale web corpora, we cannot rule out potential exposure to publicly available datasets during pretraining.
Specifically, on the seen MedNLI Semantic Entailment Generation dataset, LLAMA3-MedReadCtrl's performance substantially exceeded GPT-4 in terms of ROUGE-1 (40.27 vs. 24.68), ROUGE-L (39.49 vs. 23.90), BLEU (10.35 vs. 1.78), demonstrating its remarkable capability to generate medically accurate and semantically aligned outputs. 
On the unseen MTSamples Medical Text Simplification task, LLAMA3-MedReadCtrl notably surpassed GPT-4 by wide margins across all metrics, including ROUGE-1 (53.99 vs. 39.40), ROUGE-L (51.14 vs. 36.44), BLEU (29.73 vs. 11.63), and SARI (23.21 vs. 17.03). 
This consistently superior performance on unseen medical data demonstrates LLAMA3-MedReadCtrl’s robustness and its effectiveness in generalizing readability simplification within unseen medical contexts, which is essential for real-world clinical utility.
For general domain tasks, LLAMA3-MedReadCtrl maintained similarly robust performance. 
In the ASSET Text Simplification dataset, LLAMA3-MedReadCtrl again led the evaluation with clear margins on ROUGE-1 (66.09), ROUGE-L (59.16), BLEU (38.56), and SARI (45.60), highlighting its strong capacity for readability control in general language scenarios. 
Additionally, LLAMA3-MedReadCtrl demonstrated excellent paraphrasing skills on the PAWS dataset, achieving remarkably superior performance over GPT-4, Claude-3, and GPT-3.5 on all key metrics including ROUGE-1 (93.09 vs. GPT-4's 75.37), ROUGE-L (84.34 vs. 70.64), BLEU (66.49 significantly higher than GPT-4’s 31.22), and SARI (60.53 vs. GPT-4’s 34.35), emphasizing LLAMA3-MedReadCtrl's outstanding ability to generate precise paraphrased texts that effectively balance lexical novelty with content preservation.

\begin{table}[!htbp]
    \centering
    \scalebox{0.95}{
    \begin{tabular}{ll|cccc|cccc}
    \toprule
    & & \multicolumn{4}{c|}{\textbf{GPT-4}} & \multicolumn{4}{c}{\textbf{Ours (LLAMA3-MedReadCtrl)}} \\
    \cmidrule(lr){3-6} \cmidrule(lr){7-10}
    Dataset & Grade & Accuracy & Clarity & Consistency & Fluency & Accuracy & Clarity & Consistency & Fluency \\
    \midrule
    readme & grade 2 & 4.10 & 2.60 & 4.05 & 3.95 & \textcolor{DeepGreen}{4.25} & \textcolor{DeepGreen}{4.05} & \underline{4.05} & \textcolor{DeepGreen}{4.40} \\
    readme & grade 5 & 4.20 & 3.95 & 4.20 & 4.20 & \textcolor{DeepGreen}{4.80} & \textcolor{DeepGreen}{4.25} & \textcolor{DeepGreen}{4.80} & \textcolor{DeepGreen}{4.45} \\
    readme & grade 8 & 4.60 & 4.40 & 4.60 & 4.60 & \textcolor{DeepGreen}{4.90} & \textcolor{DeepGreen}{4.75} & \textcolor{DeepGreen}{4.90} & \textcolor{DeepGreen}{4.65} \\
    readme & grade 11 & 4.90 & 4.55 & 4.90 & 4.65 & \underline{4.90} & \textcolor{DeepGreen}{4.65} & \textcolor{red}{4.85} & \textcolor{red}{4.50} \\[1ex]
    
    mtsamples & grade 2 & 3.90 & 3.10 & 4.05 & 3.95 & \textcolor{DeepGreen}{4.20} & \textcolor{DeepGreen}{4.75} & \textcolor{DeepGreen}{4.45} & \textcolor{DeepGreen}{4.80} \\
    mtsamples & grade 5 & 3.80 & 3.80 & 3.95 & 4.00 & \textcolor{DeepGreen}{4.20} & \textcolor{DeepGreen}{4.75} & \textcolor{DeepGreen}{4.45} & \textcolor{DeepGreen}{4.85} \\
    mtsamples & grade 8 & 4.25 & 4.25 & 4.25 & 4.20 & \textcolor{DeepGreen}{4.60} & \textcolor{DeepGreen}{4.80} & \textcolor{DeepGreen}{4.60} & \textcolor{DeepGreen}{4.80} \\
    mtsamples & grade 11 & 4.35 & 4.25 & 4.35 & 4.30 & \underline{4.35} & \textcolor{DeepGreen}{4.80} & \textcolor{red}{4.30} & \textcolor{DeepGreen}{4.40} \\[1ex]
    
    mednli & grade 2 & 4.40 & 3.05 & 4.30 & 3.75 & \textcolor{DeepGreen}{4.55} & \textcolor{DeepGreen}{4.00} & \textcolor{DeepGreen}{4.40} & \textcolor{DeepGreen}{4.30} \\
    mednli & grade 5 & 4.50 & 4.40 & 4.15 & 4.00 & \textcolor{DeepGreen}{4.70} & \textcolor{DeepGreen}{4.55} & \textcolor{DeepGreen}{4.55} & \textcolor{DeepGreen}{4.30} \\
    mednli & grade 8 & 4.40 & 4.40 & 4.30 & 4.30 & \textcolor{DeepGreen}{4.90} & \textcolor{DeepGreen}{4.65} & \textcolor{DeepGreen}{4.70} & \textcolor{DeepGreen}{4.70} \\
    mednli & grade 11 & 4.50 & 4.10 & 4.50 & 3.95 & \textcolor{DeepGreen}{4.85} & \textcolor{DeepGreen}{4.70} & \textcolor{DeepGreen}{4.75} & \textcolor{DeepGreen}{4.55} \\
    \midrule

    \multicolumn{2}{c|}{\textbf{Average by Dataset}} & \multicolumn{4}{c|}{\textbf{GPT-4: mean (95\% CI)}} & \multicolumn{4}{c}{\textbf{Ours: mean (95\% CI)}} \\
    \midrule
    \multicolumn{2}{l|}{MedNLI} & \multicolumn{4}{c|}{4.19 (4.06, 4.32)} & \multicolumn{4}{c}{\textcolor{DeepGreen}{4.57 (4.49, 4.66)}} \\
    \multicolumn{2}{l|}{MTSamples} & \multicolumn{4}{c|}{4.05 (3.94, 4.16)} & \multicolumn{4}{c}{\textcolor{DeepGreen}{4.55 (4.44, 4.66)}} \\
    \multicolumn{2}{l|}{ReadMe} & \multicolumn{4}{c|}{4.28 (4.16, 4.40)} & \multicolumn{4}{c}{\textcolor{DeepGreen}{4.57 (4.48, 4.67)}} \\
    \midrule
    \multicolumn{2}{c|}{\textbf{Average by Grade}} & \multicolumn{4}{c|}{\textbf{GPT-4}} & \multicolumn{4}{c}{\textbf{Ours}} \\
    \midrule
    \multicolumn{2}{l|}{Grade 2} & \multicolumn{4}{c|}{3.77 (3.60, 3.93)} & \multicolumn{4}{c}{\textcolor{DeepGreen}{4.33 (4.19, 4.46)}} \\
    \multicolumn{2}{l|}{Grade 5} & \multicolumn{4}{c|}{4.10 (3.97, 4.22)} & \multicolumn{4}{c}{\textcolor{DeepGreen}{4.55 (4.44, 4.66)}} \\
    \multicolumn{2}{l|}{Grade 8} & \multicolumn{4}{c|}{4.38 (4.26, 4.50)} & \multicolumn{4}{c}{\textcolor{DeepGreen}{4.75 (4.66, 4.84)}} \\
    \multicolumn{2}{l|}{Grade 11} & \multicolumn{4}{c|}{4.44 (4.32, 4.56)} & \multicolumn{4}{c}{\textcolor{DeepGreen}{4.63 (4.53, 4.74)}} \\
    \midrule
    
    \multicolumn{2}{c|}{\textbf{Average by Criterion}} & \multicolumn{4}{c|}{\textbf{GPT-4}} & \multicolumn{4}{c}{\textbf{Ours}} \\
    \midrule
    \multicolumn{2}{l|}{Accuracy} & \multicolumn{4}{c|}{4.33 (4.20, 4.45)} & \multicolumn{4}{c}{\textcolor{DeepGreen}{4.61 (4.49, 4.73)}} \\
    \multicolumn{2}{l|}{Clarity} & \multicolumn{4}{c|}{3.87 (3.70, 4.04)} & \multicolumn{4}{c}{\textcolor{DeepGreen}{4.52 (4.41, 4.63)}} \\
    \multicolumn{2}{l|}{Consistency} & \multicolumn{4}{c|}{4.32 (4.20, 4.44)} & \multicolumn{4}{c}{\textcolor{DeepGreen}{4.57 (4.45, 4.69)}} \\
    \multicolumn{2}{l|}{Fluency} & \multicolumn{4}{c|}{4.17 (4.03, 4.30)} & \multicolumn{4}{c}{\textcolor{DeepGreen}{4.56 (4.46, 4.66)}} \\
    \midrule

    \multicolumn{2}{l|}{\textbf{Overall Average}} & \multicolumn{4}{c|}{4.17 (4.10, 4.24)} & \multicolumn{4}{c}{\textcolor{DeepGreen}{4.56 (4.51, 4.62)}} \\
    \bottomrule
    \end{tabular}
    }
    \caption{Human evaluation results comparing our \textbf{LLAMA3-MedReadCtrl} model and GPT-4 across biomedical datasets (MedNLI, MTSamples, ReadMe) and reading levels (Grades 2, 5, 8, 11). Evaluation dimensions include Accuracy, Clarity, Consistency, and Fluency. \textcolor{DeepGreen}{Green} means "Ours better"; \textcolor{red}{Red} means "GPT-4 better"; \underline{Underline} means "Same".}
    \label{tab:human_eval}
\end{table}

\begin{figure*}[ht!]
    \centering

    \begin{subfigure}{0.49\textwidth}
        \centering
        \includegraphics[width=\textwidth]{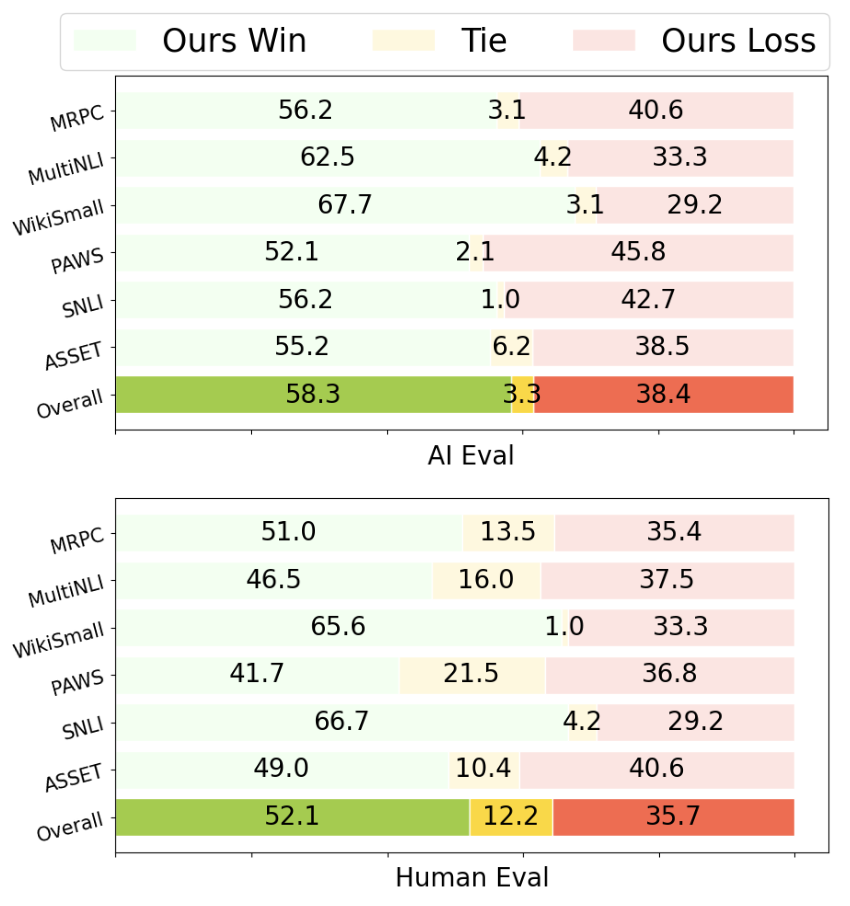}
        \caption{Ours vs GPT4 on the general domain.}
        \label{subfig:general_eval}
    \end{subfigure}
    \hfill
    \begin{subfigure}{0.49\textwidth}
        \centering
        \includegraphics[width=\textwidth]{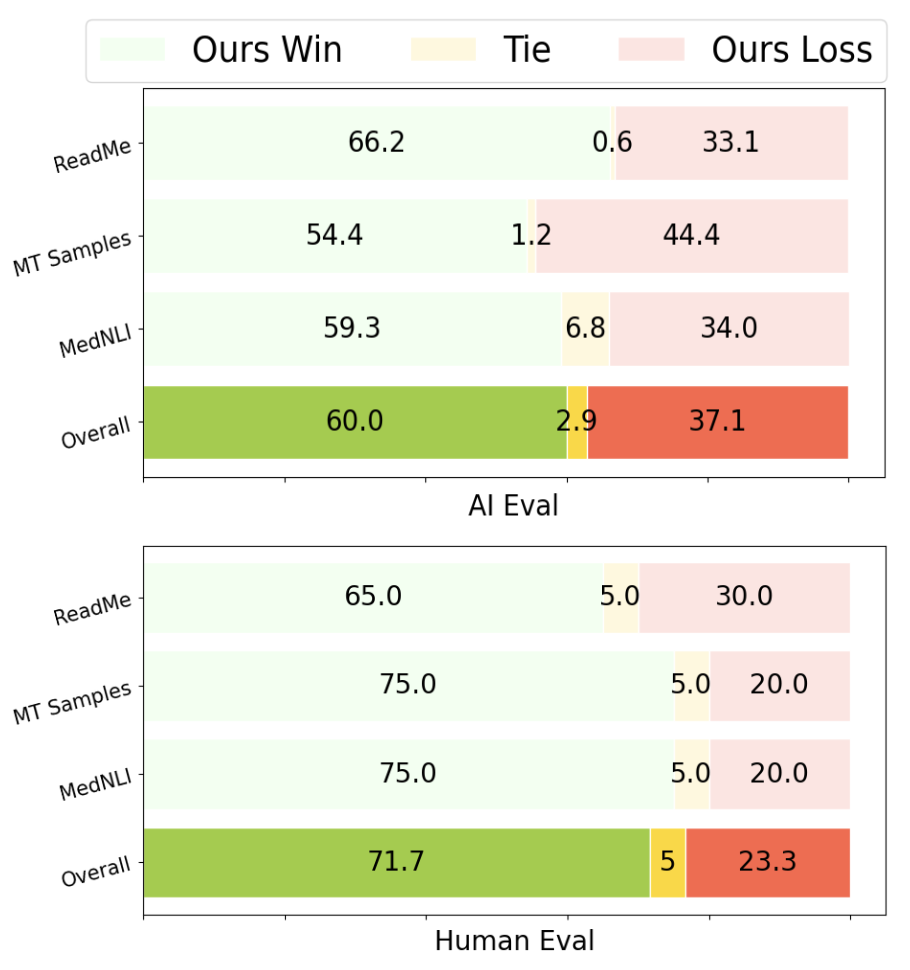}
        \caption{Ours vs GPT4 on the medical domain.}
        \label{subfig:medical_eval}
    \end{subfigure}

    \vspace{2mm}
    \caption{Human preference. Cohen's Kappa for two medical experts: 0.807.}
    \label{fig:human_ai_preference}
\end{figure*}

\subsection{Human evaluation results}

Table~\ref{tab:human_eval} presents comprehensive human evaluation results comparing LLAMA3-MedReadCtrl with GPT-4 across three biomedical datasets (MedNLI, MTSamples, ReadMe) and four readability levels (Grades 2, 5, 8, and 11), evaluated along four dimensions: Accuracy, Clarity, Consistency, and Fluency. Overall, LLAMA3-MedReadCtrl achieves superior performance, with an average score of 4.56 compared to GPT-4's 4.17. Notably, LLAMA3-MedReadCtrl demonstrates particularly strong gains at lower readability levels, where simplifying biomedical content is most challenging. For Grade 2 outputs, our model significantly outperforms GPT-4 in Clarity across all datasets—ReadMe (4.05 vs. 2.60), MTSamples (4.75 vs. 3.10), and MedNLI (4.00 vs. 3.05)—while simultaneously improving Accuracy and Consistency, indicating that readability control does not compromise content fidelity. Similar trends are observed at Grade 5, where LLAMA3-MedReadCtrl consistently achieves higher scores, with improvements of up to +0.8 points in Accuracy and Fluency. Even at higher reading levels (Grades 8 and 11), LLAMA3-MedReadCtrl maintains parity or achieves marginal gains over GPT-4, highlighting its ability to generalize across complexity levels. When broken down by evaluation criteria, LLAMA3-MedReadCtrl consistently surpasses GPT-4: its average Clarity score improves markedly from 3.87 to 4.52, while Accuracy (4.61 vs. 4.33), Consistency (4.57 vs. 4.32), and Fluency (4.56 vs. 4.17) also exhibit consistent gains. Examining individual datasets further confirms this trend; for example, on ReadMe at Grade 11, LLAMA3-MedReadCtrl achieves higher Clarity (4.65 vs. 4.55) and Fluency (4.80 vs. 4.65), and similar improvements are evident on MTSamples and MedNLI. These results collectively underscore LLAMA3-MedReadCtrl’s effectiveness in generating readability-controlled biomedical text that not only enhances simplicity but also preserves factual accuracy, logical consistency, and linguistic fluency, outperforming GPT-4 across diverse datasets and readability levels.

Figure~\ref{fig:human_ai_preference} presents human preference evaluations comparing LLAMA3-MedReadCtrl and GPT-4 across both general and medical domains, assessed by AI judges and human experts. Overall, LLAMA3-MedReadCtrl demonstrates a clear advantage in both domains. In the general domain, AI evaluations favored LLAMA3-MedReadCtrl in 58.3\% of cases, significantly higher than GPT-4’s 38.4\% preference rate. Human evaluators exhibit a consistent trend, preferring LLAMA3-MedReadCtrl in 52.1\% of cases compared to GPT-4’s 35.7\%. Notably, datasets such as WikiSmall and SNLI show particularly strong human preference for LLAMA3-MedReadCtrl, reaching 65.6\% and 66.7\% respectively, underscoring its robustness in tasks like text simplification and natural language inference.
In the medical domain, LLAMA3-MedReadCtrl’s superiority is even more pronounced. AI judges preferred our model in 60.0\% of cases versus GPT-4’s 37.1\%. More importantly, human expert evaluations further amplify this difference: LLAMA3-MedReadCtrl achieves a striking overall preference rate of 71.7\%, compared to just 23.3\% for GPT-4. Expert agreement is remarkably high, with a Cohen’s Kappa of 0.807, indicating strong annotation reliability. Examining individual datasets reveals consistent advantages—LLAMA3-MedReadCtrl achieves over 65\% expert preference on ReadMe, MTSamples, and MedNLI, with both MTSamples and MedNLI reaching 75.0\%. These results not only reaffirm LLAMA3-MedReadCtrl’s effectiveness in medical domains but also highlight its ability to generate text that aligns closely with expert expectations in terms of readability, factual accuracy, and overall quality, making it particularly well-suited for high-stakes applications such as patient education and clinical communication.

\section{Discussion}
\label{discussion}

Effective communication in healthcare hinges not only on clinical accuracy but also on the ability to convey information in a manner aligned with patients’ literacy levels and comprehension needs~\cite{sauerbrei2023impact,robinson2024reviewing}.
Our results demonstrate that LLAMA3-MedReadCtrl meaningfully advances this goal by enabling large language models to dynamically control text readability while preserving content quality and relevance—a capability critical for patient education, informed consent, and shared decision-making~\cite{rajpurkar2022ai,silcox2024potential}.
Compared to leading proprietary models such as GPT-4 and Claude-3, as well as the baseline LLAMA3 model, MedReadCtrl achieves superior readability control across both medical and general domains (Figure 1), with particularly strong performance on clinical datasets where simplification must be balanced with medical precision. 
Automatic metrics (Table 1) further support this: MedReadCtrl consistently yields higher ROUGE, BLEU, and SARI scores across diverse tasks and unseen datasets, demonstrating both robustness and generalizability.

Human evaluations add an important layer of interpretability. At lower grade levels—especially Grades 2 and 5, where health literacy challenges are most acute—MedReadCtrl outperforms GPT-4 in clarity, fluency, and clinical accuracy (Table 2, Figure 2). 
These improvements are more than stylistic; they translate directly into better comprehension for patients who may otherwise misinterpret discharge instructions, medication labels, or follow-up plans \cite{stanceski2024quality}.
Notably, MedReadCtrl maintains semantic fidelity even when simplifying technical content, avoiding jargon while preserving intent—an essential property for equitable, patient-centered care delivery \cite{nutbeam2023artificial}.

At lower readability levels, the ability to convey biomedical information in accessible and age-appropriate language is essential for reducing comprehension disparities among low-literacy populations, including children, older adults, and non-native speakers~\cite{vo2023association,morrison2019health,suen2025association}.
These groups face heightened risk of misunderstanding key health information, which can contribute to poor treatment adherence, medication errors, or delayed care-seeking behavior. 
In this context, LLAMA3-MedReadCtrl demonstrates strong alignment with health literacy guidelines by consistently avoiding complex or technical terminology and producing intuitive, patient-friendly explanations. 
For instance, in the Grade 2 example describing an X-ray procedure (Table~\ref{tab:qualitative_analysis_low_read}), GPT-4 generates: “Taking an X-ray picture of the spinal cord after putting special dye into a space around it.” 
While structurally simplified, it retains specialized vocabulary such as “X-ray” and “dye,” which may be unfamiliar to early readers.
In contrast, MedReadCtrl rewrites this sentence as: “A special picture of the spine that helps doctors see inside,” eliminating jargon while preserving the medical intent.
This form of semantic simplification aligns with plain language principles and supports comprehension among young patients or caregivers with limited background knowledge \cite{achiam2023gpt}.
A similar pattern appears in the case describing a cough with sputum and blood streaks. GPT-4’s Grade 2 output—“She has a cough with some phlegm and a little bit of blood sometimes, but no big blood”—includes opaque terms like “phlegm” and uses awkward quantifiers that disrupt fluency. 
MedReadCtrl instead produces: “She has a cough and sometimes brings up yucky stuff from her lungs. Sometimes it might have a little bit of red in it, but it’s not too much,” employing concrete, relatable language that mirrors health educator strategies in pediatric and low-literacy settings. Another compelling example involves anatomical description at Grade 5. 
GPT-4’s sentence—“She does have some soreness in her groin on both sides”—retains the anatomically correct but potentially confusing term “groin.” 
MedReadCtrl transforms this into: “She does have some pain in the area where her legs and hips meet, on both sides,” offering a precise yet accessible explanation that would better support health comprehension in community education portals, post-visit instructions, or telehealth consultations \cite{nutbeam2023artificial}. 
These adaptations reflect the model’s capacity to restructure content in a way that prioritizes understanding over literal term retention—mirroring strategies used by professional health educators and offering critical value for patient-centered communication in post-visit instructions, pediatric care, and health literacy tools \cite{ahrq2024toolkit}.

At higher readability levels (Grades 8 and 11), MedReadCtrl continues to offer value through more context-sensitive and humanized phrasing. 
For example, in the Grade 8 simplification of “depression treatment drug” (Table~\ref{tab:qualitative_analysis_high_read}), GPT-4 generates the generic phrase “a medicine for treating depression,” which is technically accurate but semantically flat. 
In contrast, MedReadCtrl rewrites this as: “A medicine that helps people feel better when they are sad,” offering a more empathetic and humanized description. 
Similarly, in a Grade 11 paraphrase of erectile dysfunction, GPT-4 states: “The inability of a man to get an erection because of mental issues or problems with his body.” 
MedReadCtrl’s version—“A condition in men where they are unable to get or maintain an erection, often due to a physical or mental health issue”—uses more clinical but naturalistic phrasing and avoids stigmatizing language, which may be especially important for sensitive topics. 
This level of semantic elaboration can support communication not only with patients but also with caregivers, educators, and multidisciplinary teams involved in long-term condition management \cite{schooley2020integrated,stephen2024staff}.
A particularly illustrative demonstration of MedReadCtrl’s readability control occurs in the context of follow-up care planning. 
GPT-4 generates the same sentence—“I will check back in three months”—across all readability levels, showing no sensitivity to audience comprehension needs. 
In contrast, MedReadCtrl scales the expression appropriately: for Grade 2, it outputs “I will see you again in three months,” using simple, conversational phrasing. 
By Grade 11, the model elaborates to: “I will schedule a follow-up appointment with you in approximately three months to assess the status of your condition and determine any necessary adjustments to your treatment plan.”
This progression illustrates the model’s ability to integrate both linguistic complexity and communicative intent—an essential capability for generating documentation templates and educational materials tailored to adolescents, caregivers, or patients managing long-term conditions \cite{nutbeam2023artificial}.

Taken together, these examples underscore MedReadCtrl’s strength in generating audience-appropriate, semantically faithful text across a range of literacy levels. 
The model goes beyond surface simplification by restructuring sentence syntax, reframing technical terms, and adjusting tone—all while maintaining clinical accuracy. 
These capabilities are essential for patient-facing technologies, such as discharge instruction portals, caregiver education tools, and AI-powered chatbots, which must adapt content to diverse comprehension needs in real-world settings \cite{sauerbrei2023impact,robinson2024reviewing,nutbeam2023artificial}.

Despite these strengths, several limitations remain. First, like other large language models, MedReadCtrl is susceptible to factual hallucinations. 
As shown in Table~\ref{tab:error_analysis} (Case 1), the model occasionally introduces subtle inaccuracies that may mislead clinical interpretation, such as incorrectly attributing throat dryness to a symptom rather than to a condition. 
In another example, the phrase “increasing chest pain episodes” is paraphrased as “he gets hurt in his chest a lot,” which incorrectly frames angina-like symptoms as physical injury.
While seemingly minor, such distortions can erode trust or lead to serious miscommunication in diagnostic explanations and treatment decisions, especially in patient self-management or telemedicine settings. 
Integration with retrieval-augmented generation (RAG) frameworks~\cite{lewis2020retrieval} or structured biomedical knowledge sources (e.g., medical dictionaries or biomedical knowledge graphs) may help reduce the risk of hallucination and improve factual grounding.
Additionally, incorporating fact-checking frameworks~\cite{chung2025verifact} to validate outputs against trusted references before generation may improve accuracy in sensitive clinical scenarios.

Second, we observe that readability control sometimes becomes conflated with verbosity, particularly at higher readability levels. 
In Case 2, the model’s Grade 11 explanation of Sjögren’s syndrome becomes overly technical, introducing complex constructs such as “autoimmune destruction of salivary glands,” along with redundant phrases like “the second form is secondary.” 
These choices can overwhelm target readers, including adolescents or adult learners. Future iterations could incorporate multidimensional control signals—encoding not only readability grade but also length, tone, or reader type—within the prompt structure~\cite{yuan2024following}. 
Reinforcement learning with human feedback (RLHF)~\cite{jie2023prompt}, especially when combined with preference models that penalize redundant or overly medicalized outputs, may further refine generation.
Incorporating examples with graded prompts (“explain to a high schooler in 2 sentences”) during fine-tuning may improve precision in stylistic control.
In future iterations, human-annotated readability and stylistic preference healthcare data can inform reward modeling under RLHF frameworks, enabling more precise alignment with user expectations and literacy needs.

A third challenge concerns the occasional misinterpretation of instruction. As illustrated in Case 3, when presented with the input "Description: A sample note on Rheumatoid Arthritis," the model incorrectly generates a definition rather than simplifying the content.
This indicates a failure to differentiate between content-bearing inputs and meta-instructions, a weakness exacerbated in zero- or low-shot generalization scenarios. 
Mitigating this issue may require explicit instruction boundary encoding or task-specific tokens that delimit transformations versus descriptions.
Fine-tuning on synthetic instruction-following data with varied structure, along with task-grounded prompt engineering strategies~\cite{zhang2023instruction, qin2024infobench}, may also improve adherence.

Additionally, discrepancies between expert judgment and automated LLM-as-Judge assessments raise concerns about current evaluation methods. 
Although MedReadCtrl received a 71.6\% preference score from human clinicians, it received only a 60.0\% preference score from LLM-based judges. 
This divergence suggests that existing automatic evaluators may undercapture clinical relevance, semantic fidelity, and subtle differences in readability optimization.
For instance, in a Grade 5 simplification task, clinicians preferred MedReadCtrl's empathetic phrasing—“pain in the area where the legs and hips meet”—as it demonstrated improved contextual sensitivity and accessibility for lay audiences. In contrast, the LLM-as-Judge assigned a higher score to GPT-4’s output—“groin pain”—which, while medically accurate, retained potentially unfamiliar terminology. 
This illustrates the current evaluators’ blind spots when judging language adapted for low-literacy or sensitive contexts. Particularly in domains such as pediatric or mental health communication, black-box metrics may be insufficient.
Moreover, limitations such as evaluation bias, domain contamination, and interpretability issues have been documented in prior work~\cite{gu2024survey, szymanski2024limitations, li2025preference}.
Future benchmarks should consider hybrid evaluation pipelines that combine expert-derived annotations (e.g., health literacy compliance checklists or comprehension probes) with LLM-based assessments, enabling more nuanced and trustworthy evaluations.

In sum, MedReadCtrl demonstrates that fine-grained readability control is both technically feasible and clinically valuable. 
By enabling personalized content generation tailored to varying health literacy levels, the system addresses longstanding gaps in patient-centered communication and equitable access to medical information \cite{nutbeam2023artificial}.
Realizing its full potential will require continued attention to hallucination mitigation, stylistic calibration, task clarity, and robust evaluation frameworks.
As large language models are increasingly deployed in clinical education tools, caregiver interfaces, and remote health portals, systems like MedReadCtrl can serve as foundational infrastructure for safe, comprehensible, and inclusive medical AI communication~\cite{aydin2024large}.
To ensure trustworthy and scalable deployment in real-world settings, future work should also consider model transparency, regulatory compliance, and human-in-the-loop oversight strategies \cite{panch2019artificial,dennstadt2025implementing}. 
Such efforts can not only improve system reliability but also inform the development of policy and governance frameworks for responsible LLM integration into patient-facing applications~\cite{bouderhem2024shaping}.

While our findings establish MedReadCtrl’s effectiveness in producing semantically faithful, audience-appropriate outputs across diverse readability levels, several limitations warrant consideration. 
First, all experiments were conducted on English-language datasets, limiting the generalizability of our findings to non-English settings. Extending this framework to support multilingual or culturally adapted readability control remains an important direction for future research. 
Second, our study focused on three NLG tasks—text simplification, paraphrase generation, and semantic entailment generation—which, while representative, do not encompass the full range of document styles and discourse types seen in clinical communication (e.g., patient-provider dialogues~\cite{cai2023paniniqa}).
Further work is needed to explore how readability control operates across broader clinical tasks and note types.
Additionally, our use of traditional readability metrics, while widely adopted in the literature, may be insufficient to assess fluency and understandability in fragmented or telegraphic clinical texts, such as EHRs. 
Alternative approaches—such as domain-adapted, machine learning–based readability scoring—may provide more robust assessment. 
Lastly, while our human evaluation included expert scoring across multiple grade levels, it did not incorporate real users (e.g., patients or caregivers) from the target literacy groups.
Future work should include end-user evaluation to assess practical effectiveness and refine feedback loops in human-centered settings. 
These limitations notwithstanding, our work provides a scalable and adaptable foundation for personalized medical text generation, helping bridge the persistent health literacy gap in clinical care delivery.

\section{Methods}
\label{methods}

\subsection{Task Overview}

Our methodology is designed to evaluate the effectiveness of instruction tuning conditional on readability across a suite of tasks, specifically focusing on text simplifications, paraphrase generation, and semantic entailment generation. 
These tasks are strategically chosen to test the model's capability in adjusting the complexity of its output to match specified readability levels. 
They serve a broad spectrum of applications, from enhancing educational material accessibility to refining technical documentation for diverse audiences.

\begin{itemize}
[leftmargin=.1in,topsep=0.3pt]
\setlength\itemsep{0em}
\vspace{-0.2em}
    \item \textbf{Text Simplifications:} Here, the aim is to reduce the readability level of the given input text, making it more accessible to a wider audience or readers with varying comprehension skills. This task challenges the model to simplify complex text while preserving its essential content and meaning, demonstrating the ability to decrease textual complexity upon demand.
        
    \item \textbf{Paraphrase Generation:} In this task, the model is tasked with rewording the given text to produce a paraphrase that maintains the original's readability level. This requires a nuanced understanding of language to ensure the output remains true to the input's complexity and style, facilitating content reformulation without altering its accessibility.
    
    \item \textbf{Semantic Entailment Generation:} This involves creating text that semantically follows from the given input, with the flexibility to increase or decrease the readability level. The model must grasp the underlying meaning of the input text and generate output that logically entails the input, demonstrating versatility in producing content with adjustable complexity levels.

\end{itemize}

\begin{figure}[!ht]
    \centering
    \includegraphics[width=1.0\textwidth]{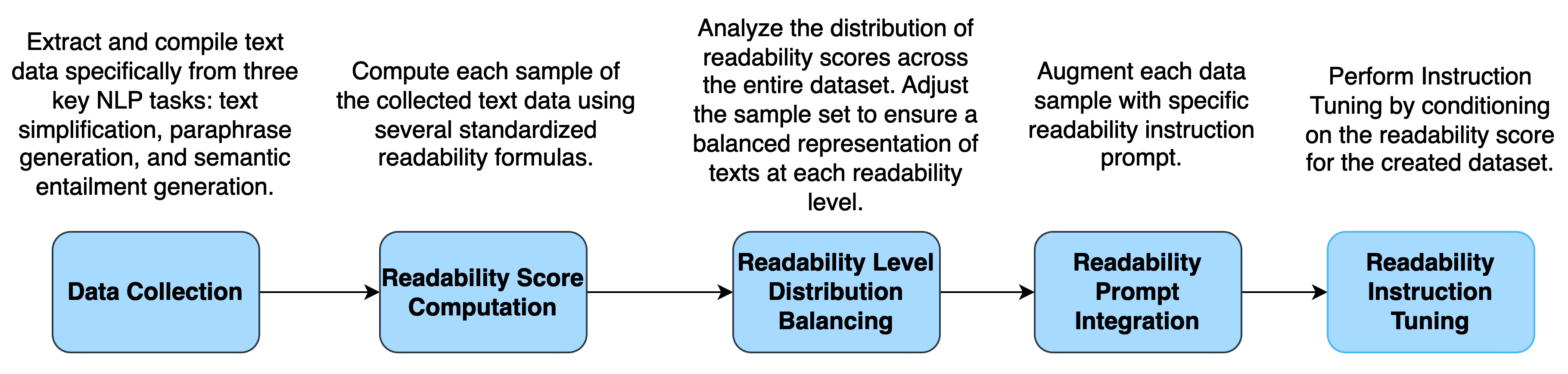}
    \caption{Overview of MedReadCtrl data construction.}
    \label{fig:method_overview}
\end{figure}

We employ the instruction tuning approach conditional on readability for all these tasks. 
This method provides explicit instructions to the model to control the output text's readability score, ensuring that the generated content aligns with the intended complexity level for the target audience.
This approach underlines our belief that these tasks can all contribute to readability control generation, where, depending on the task—be it text simplification, paraphrase generation, or semantic entailment generation—the model is calibrated to generate output with the desired readability level. 
In text simplification, the goal is to lower the readability of the output relative to the input, while in paraphrase generation, the output's readability should mirror the input's. 
For the semantic entailment generation task, the output's readability may vary, being either higher or lower than the input's, thereby offering a versatile tool for adjusting text complexity across a wide range of contexts.

\subsection{Instruction Design for Readability Control Instruction Learning}
Instruction Tuning enhances LLMs by training them on NLP tasks formatted as natural language instructions, enabling models to generate structured responses \cite{wei2021finetuned}. 
This technique has been used to transform English-centric LLMs into open-ended chat models with GPT-like performance by leveraging distilled data from GPT itself \cite{wang2022self}. 
Our work follows a similar approach, applying instruction tuning to LLM using our created dataset in Section~\ref{dataset}.
Unlike traditional supervised fine-tuning (SFT), which relies on manual annotation or synthetic data for readability control, instruction tuning provides a more generalizable approach \cite{zhang2023instruction}. It enables models to leverage their pre-trained knowledge while learning how to follow structured instructions effectively. 
Recent advances \cite{wei2021finetuned,wang2022self,tran2023bioinstruct} highlight the benefits of instruction learning for adaptability across tasks.
In our study, we adopt a FLAN-style instruction fine-tuning method \cite{wei2021finetuned} to train models on task-specific instructions for MedReadCtrl. This approach enhances LLMs' ability to follow readability-controlled instructions, improving usability while reducing the resource-intensive demands of domain-specific fine-tuning.

To achieve the desired readability level across various tasks, we employ straightforward and singular instruction. This approach emphasizes the model's ability to tailor its output to meet specific readability goals, demonstrating its versatility and effectiveness in readability control. The instruction is as follows: ``\emph{Given an input text, please output an entailment with a readability score around {target readability score}.}
This concise instruction mandates the model to generate content that not only semantically follows from the given input but also aligns with a specified readability level, showcasing the model's capacity to produce targeted outputs that cater to diverse comprehension needs and preferences.

\subsection*{Implementation and Readability Scoring}

The readability of the generated text is quantitatively evaluated using a suite of established readability metrics. We calculate the following readability scores:

\begin{itemize}[leftmargin=.1in,topsep=0.3pt]
\setlength\itemsep{0em}
    \item \textbf{Flesch-Kincaid Grade Level (FKGL)}: Estimates the U.S. grade level required for comprehension, calculated as:
    \begin{equation}
        \text{FKGL} = 206.835 - 1.015 \left(\frac{\text{totalWords}}{\text{totalSentences}}\right) - 84.6 \left(\frac{\text{totalSyllables}}{\text{totalWords}}\right).
    \end{equation}
    \item \textbf{Gunning Fog Index (GFI)}: Measures the years of formal education needed for first-read comprehension, given by:
    \begin{equation}
        GFI = 0.4 \left( \frac{\text{totalWords}}{\text{totalSentences}} + 100 \frac{\text{longWords}}{\text{totalWords}} \right),
    \end{equation}
    where \textit{longWords} are words with more than seven characters.
    \item \textbf{Automated Readability Index (ARI)}: Correlates readability with U.S. school grade level:
    \begin{equation}
        ARI = 4.71 \left(\frac{\text{totalCharacters}}{\text{totalWords}}\right) + 0.5 \left(\frac{\text{totalWords}}{\text{totalSentences}}\right) - 21.43.
    \end{equation}
    \item \textbf{Coleman-Liau Index (CLI)}: Uses character-based analysis for readability estimation:
    \begin{equation}
        CLI = 0.0588L - 0.296S - 15.8,
    \end{equation}
    where $L$ is the average number of letters per 100 words, and $S$ is the average number of sentences per 100 words.
\end{itemize}

These metrics are selected for their diverse approaches to assessing text complexity, offering a comprehensive understanding of the text's readability. 
Subsequently, an average Reading Grade Level (RGL) is derived from these scores to represent the text's overall readability. The integration of these readability assessments into our methodology allows a nuanced approach to generating text that meets the specified readability criteria. By adjusting the instruction based on the target RGL, we can fine-tune the complexity of the output, making our approach adaptable to a wide range of applications, from educational content to technical documentation. This process underscores the importance of readability in tailoring content to specific audience needs, a critical factor in communication effectiveness across various domains.

\subsection{Dataset}
\label{dataset}

Our experimental framework is designed to assess the model's performance across various tasks, specifically focusing on text simplification, paraphrase generation, and semantic entailment generation. To facilitate a comprehensive evaluation, we utilize six distinct datasets, two for each task, which enables us to explore the model's capabilities in both seen and unseen settings. The datasets employed in our experiments are outlined as follows:

\begin{itemize}
[leftmargin=.1in,topsep=0.5pt]
\setlength\itemsep{0em}
\vspace{-0.2em}
    \item \textbf{Text Simplification:} For this task, we use the ASSET \cite{alva2020asset} and WikiSmall \cite{zhu2010monolingual} datasets. ASSET is a diverse corpus for automatic sentence simplification, providing high-quality simplifications with multiple references per source sentence, making it ideal for instruction tuning and evaluation in seen settings. WikiSmall serves as an additional dataset for evaluating performance in an unseen setting, offering a different collection of simplified sentences derived from Wikipedia articles.

    \item \textbf{Medical Text Simplification:} For this task, we utilize the ReadME and MTSamples datasets. The README dataset \cite{yao2024readme} contains extensive pairs of medical jargon~\cite{kwon2022medjex} and their corresponding lay definitions, making it ideal for training models to convert complex medical jargon into patient-friendly language in seen settings. 
    MTSamples \cite{moramarco2022towards} provides a rich collection of medical transcription reports across numerous specialties, allowing models to further generalize and evaluate simplification performance in an unseen, clinically varied setting.
    Together, these datasets support the development of systems that simplify medical language, enhancing patient comprehension and accessibility in healthcare communication.

    \item \textbf{Paraphrase Generation:} We utilize the PAWS \cite{zhang2019paws} (Paraphrase Adversaries from Word Scrambling) and MRPC (Microsoft Research Paraphrase Corpus) \cite{dolan2005automatically} datasets. PAWS contains pairs of sentences paraphrasing each other, including those constructed through controlled word scrambling, making it suitable for training and the seen setting evaluations. MRPC offers a collection of sentence pairs labeled as paraphrases or not, sourced from online news sources, to test the model's paraphrasing ability in unseen settings.

    \item \textbf{General Semantic Entailment Generation:} For this task, the SNLI (Stanford Natural Language Inference) \cite{bowman2015large} and MultiNLI (Multi-Genre Natural Language Inference) \cite{williams2017broad} datasets are employed. SNLI is a large collection of sentence pairs annotated with textual entailment information, used for instruction tuning and seen setting evaluation. MultiNLI extends this to a broader range of genres and contexts, providing a robust challenge for the model in unseen settings. Together, these datasets enable the model to learn and generalize entailment patterns across a wide range of non-specialized texts.

    \item \textbf{Medical Semantic Entailment Generation:} In the biomedical domain, we employ the MedNLI (Medical Natural Language Inference) dataset \cite{romanov2018lessons} for entailment tasks specific to clinical settings. Unlike SNLI and MultiNLI, MedNLI consists of sentence pairs derived from electronic health records (EHRs) where medical terminology, context, and clinical reasoning are critical for accurate entailment. This dataset allows the model to focus on domain-specific entailment challenges, making it especially useful for evaluating inference abilities in unseen, medically-focused settings.

\end{itemize}

In our experimental setup, instruction tuning is performed on the training sets of ASSET, PAWS, and SNLI to align the model's output with specific readability goals. The effectiveness of this approach is then evaluated in two distinct settings: a \textit{seen setting}, using the test sets of ASSET, PAWS, and SNLI, and an \textit{unseen setting}, using the test sets of WikiSmall, MRPC, and MultiNLI. This methodology allows us to not only measure the model's immediate response to the instruction tuning but also its generalizability and adaptability to different textual contexts and tasks.

\begin{table}[h]
    \centering
    \scriptsize
    \renewcommand{\arraystretch}{1.2}
    \resizebox{\textwidth}{!}{  
    \begin{tabular}{ll|cccccccccccc|c}
        \toprule
        \multicolumn{2}{c|}{\textbf{Dataset (Split)}} & \textbf{Grade 1} & \textbf{Grade 2} & \textbf{Grade 3} & \textbf{Grade 4} & \textbf{Grade 5} & \textbf{Grade 6} & \textbf{Grade 7} & \textbf{Grade 8} & \textbf{Grade 9} & \textbf{Grade 10} & \textbf{Grade 11} & \textbf{Grade 12} & \textbf{Total} \\
        \midrule
        \multirow{3}{*}{\textbf{ReadMe}} & Train  & 150(6.10\%)  & 180(7.32\%)  & 170(6.91\%)  & 160(6.50\%)  & 190(7.72\%) & 200(8.13\%) & 210(8.54\%) & 220(8.94\%) & 230(9.35\%) & 240(9.76\%) & 250(10.16\%) & 260(10.57\%) & 2460 \\
                                         & Dev    & 130(6.22\%) & 140(6.69\%) & 150(7.17\%) & 160(7.65\%) & 170(8.13\%) & 180(8.61\%) & 190(9.09\%) & 200(9.57\%) & 210(10.05\%) & 220(10.53\%) & 230(11.01\%) & 240(11.49\%) & 2090 \\
                                         & Test   & 140(6.22\%) & 150(6.67\%) & 160(7.11\%) & 170(7.56\%) & 180(8.00\%) & 190(8.44\%) & 200(8.89\%) & 210(9.33\%) & 220(9.78\%) & 230(10.22\%) & 240(10.67\%) & 250(11.11\%) & 2250 \\
        \midrule
        \multirow{3}{*}{\textbf{MedNLI}} & Train  & 160(6.36\%)  & 170(6.75\%)  & 180(7.14\%)  & 190(7.54\%)  & 200(7.94\%) & 210(8.33\%) & 220(8.73\%) & 230(9.13\%) & 240(9.52\%) & 250(9.92\%) & 260(10.32\%) & 270(10.71\%) & 2520 \\
                                         & Dev    & 140(6.42\%)  & 150(6.88\%)  & 160(7.34\%)  & 170(7.80\%)  & 180(8.26\%) & 190(8.72\%) & 200(9.17\%) & 210(9.63\%) & 220(10.09\%) & 230(10.55\%) & 240(11.01\%) & 250(11.47\%) & 2180 \\
                                         & Test   & 150(6.28\%)  & 160(6.70\%)  & 170(7.11\%)  & 180(7.53\%)  & 190(7.94\%) & 200(8.36\%) & 210(8.77\%) & 220(9.19\%) & 230(9.61\%) & 240(10.03\%) & 250(10.45\%) & 260(10.86\%) & 2390 \\
        \midrule
        \multirow{3}{*}{\textbf{Asset}}  & Train  & 170(6.50\%)  & 180(6.88\%)  & 190(7.26\%)  & 200(7.64\%)  & 210(8.02\%) & 220(8.40\%) & 230(8.78\%) & 240(9.16\%) & 250(9.54\%) & 260(9.92\%) & 270(10.30\%) & 280(10.68\%) & 2620 \\
                                         & Dev    & 150(6.12\%)  & 160(6.53\%)  & 170(6.94\%)  & 180(7.35\%)  & 190(7.76\%) & 200(8.16\%) & 210(8.57\%) & 220(8.98\%) & 230(9.39\%) & 240(9.80\%) & 250(10.20\%) & 260(10.61\%) & 2450 \\
                                         & Test   & 160(6.32\%)  & 170(6.71\%)  & 180(7.09\%)  & 190(7.47\%)  & 200(7.85\%) & 210(8.23\%) & 220(8.61\%) & 230(8.99\%) & 240(9.38\%) & 250(9.76\%) & 260(10.14\%) & 270(10.52\%) & 2550 \\
        \midrule
        \multirow{3}{*}{\textbf{SNLI}}   & Train  & 110(5.00\%)  & 120(5.50\%)  & 130(6.00\%)  & 140(6.50\%)  & 150(7.00\%) & 160(7.50\%) & 200(9.50\%) & 210(10.00\%) & 220(10.50\%) & 230(11.00\%) & 240(11.50\%) & 250(12.00\%) & 2100 \\
                                         & Dev    & 100(5.26\%)  & 110(5.79\%)  & 120(6.32\%)  & 130(6.84\%)  & 140(7.37\%) & 150(7.89\%) & 180(9.47\%) & 190(10.00\%) & 200(10.53\%) & 210(11.05\%) & 220(11.58\%) & 230(12.11\%) & 1900 \\
                                         & Test   & 110(5.08\%)  & 120(5.56\%)  & 130(6.05\%)  & 140(6.53\%)  & 150(7.02\%) & 160(7.51\%) & 190(8.91\%) & 200(9.40\%) & 210(9.89\%) & 220(10.38\%) & 230(10.87\%) & 240(11.36\%) & 2100 \\
                                         
        \midrule
        \multirow{3}{*}{\textbf{PAWS}}   & Train  & 180(6.72\%)  & 190(7.10\%)  & 200(7.46\%)  & 210(7.84\%)  & 220(8.20\%) & 230(8.56\%) & 240(8.92\%) & 250(9.28\%) & 260(9.64\%) & 270(10.00\%) & 280(10.36\%) & 290(10.72\%) & 2700 \\
                                         & Dev    & 160(6.35\%)  & 170(6.75\%)  & 180(7.14\%)  & 190(7.54\%)  & 200(7.94\%) & 210(8.33\%) & 220(8.73\%) & 230(9.13\%) & 240(9.52\%) & 250(9.92\%) & 260(10.32\%) & 270(10.71\%) & 2520 \\
                                         & Test   & 170(6.46\%)  & 180(6.84\%)  & 190(7.22\%)  & 200(7.60\%)  & 210(7.98\%) & 220(8.36\%) & 230(8.74\%) & 240(9.12\%) & 250(9.50\%) & 260(9.88\%) & 270(10.26\%) & 280(10.64\%) & 2630 \\

        \midrule
        \multirow{1}{*}{\textbf{MTSamples}} & Test  & 88(10.11\%)  & 78(8.97\%)  & 64(7.36\%)  & 92(10.57\%)  & 57(6.55\%)  & 70(8.05\%)  & 88(10.11\%)  & 68(7.82\%)  & 72(8.28\%)  & 60(6.90\%)  & 60(6.90\%)  & 73(8.39\%)  & 870 \\
         \midrule
        \multirow{1}{*}{\textbf{WikiSmall}} & Test  & 85(9.73\%)  & 89(10.18\%)  & 73(8.35\%)  & 52(5.95\%)  & 71(8.12\%)  & 51(5.84\%)  & 73(8.35\%)  & 93(10.64\%)  & 79(9.04\%)  & 87(9.95\%)  & 51(5.84\%)  & 70(8.01\%)  & 874 \\
        \midrule
        \multirow{1}{*}{\textbf{MultiNLI}} & Test  & 78(8.49\%)  &  59(6.42\%)  &  67(7.29\%)  &  90(9.80\%)  &  72(7.84\%)  &  97(10.57\%)  &  75(8.17\%)  &  89(9.71\%)  &  76(8.29\%)  &  62(6.76\%)  &  61(6.65\%)  &  94(10.25\%)  & 920 \\

        \midrule
        \multirow{1}{*}{\textbf{MRPC}} & Test  & 82(8.65\%)  &  61(6.43\%)  &  71(7.49\%)  &  93(9.81\%)  &  74(7.81\%)  &  98(10.34\%)  &  76(8.02\%)  &  91(9.60\%)  &  77(8.12\%)  &  65(6.86\%)  &  64(6.75\%)  &  96(10.13\%)  & 948 \\

        \bottomrule
    \end{tabular}
    }
    \caption{Distribution of examples readability scores from instruction tuning datasets}
    \label{tab:readability_distribution}
\end{table}

\subsection{Evaluation Metrics}

To comprehensively evaluate the model's performance across the different tasks, we employ a multifaceted set of metrics that assess various aspects of the generated texts. These metrics enable us to gauge the model's effectiveness in adjusting readability, maintaining factual accuracy, and ensuring textual coherence and consistency. The following metrics are used:

\begin{itemize}
[leftmargin=.1in,topsep=0.3pt]
\setlength\itemsep{0em}
\vspace{-0.2em}
    \item \textbf{ROUGE:} The ROUGE (Recall-Oriented Understudy for Gisting Evaluation) \cite{lin2004rouge} metric is used to evaluate the quality of text simplification and summarization tasks. Specifically, we employ ROUGE-1 and ROUGE-L, which measure the overlap of unigrams and the longest common subsequence between the generated texts and reference texts, respectively. ROUGE-1 captures basic content similarity, while ROUGE-L assesses the fluency and coherence of the generated text, providing a comprehensive evaluation of the model’s ability to retain key information while simplifying or summarizing content.

    \item \textbf{BLEU:} The BLEU (Bilingual Evaluation Understudy) \cite{papineni2002bleu} metric is applied to evaluate paraphrase generation and semantic entailment tasks. It quantifies the linguistic similarity between the generated texts and reference texts, indicating the model's capability to produce coherent and contextually appropriate content.

    \item \textbf{SARI:} The SARI (System output Against References and the Input sentence) \cite{xu2016optimizing} metric is utilized to assess the quality of text simplification. It measures the model's ability to produce simplified text that is both accurate and helpful, comparing the generated output against both the original text and reference simplifications.
    
\end{itemize}

These metrics collectively offer a robust framework for assessing the nuanced performance of the model across various dimensions of text generation, readability adjustment, and content quality.

\subsection{Evaluated Models}
In our study, we evaluate a diverse set of models to understand their efficacy in handling tasks related to term definition generation, text simplification, and text complication, particularly focusing on adjusting text complexity according to specified readability levels. The models include:

\begin{itemize}
[leftmargin=.1in,topsep=0.3pt]
\setlength\itemsep{0em}
\vspace{-0.2em}
    \item \textbf{GPT-3.5:} As a precursor to GPT-4, GPT-3.5 has demonstrated substantial capabilities in generating human-like text across various tasks. It serves as a baseline to understand the incremental improvements brought about by its successors and other models.
    
    \item \textbf{GPT-4:} The latest iteration from OpenAI's GPT series at the time of our study, GPT-4, represents a significant leap in language model performance, offering improved comprehension and generation capabilities over its predecessors.
    
    \item \textbf{Claude-3:} As a model known for its understanding and generation abilities, Claude-3 has been included as a baseline for its efficiency in handling various NLP tasks and its purported adaptability to instruction-based prompts, making it a relevant comparison for our instruction-tuned model.
    
    \item \textbf{LLaMA3 8B MedReadCtrl:} Our proposed model has been instruction-tuned to adjust the readability level of generated texts based on explicit instructions. LLaMA3 8B is designed to excel in the specific tasks of text simplification, paraphrase generation, and semantic entailment generation, leveraging instruction tuning to achieve precise control over the readability of its outputs.
\end{itemize}

Each of these models brings unique strengths and capabilities to the table, allowing us to conduct a comprehensive comparison that not only highlights LLaMA3 8B's advancements in controlling readability but also situates these achievements within the broader context of current NLP technologies. By evaluating LLaMA3 8B against these established models, we aim to demonstrate its efficacy and potential applications in enhancing readability control in automatic text generation.

\subsection{Hyper-parameter Settings}
The experiments were executed using the version 4.37.1 of the transformers library released by Hugging Face. In Table~\ref{tab:hyperparameters}, we report the hyperparameters used to train the models on our combined dataset. We use the Adam optimizer and employ a linearly decreasing learning rate schedule with warm-up step is 200. In this section, we detail our experimental setup, the datasets employed, and the evaluation strategy adopted for assessing the performance of our instruction-tuned LLMs in various BioNLP tasks. Furthermore, all experiments were conducted using two Nvidia A100 GPUs, each with 40 GB of memory. The CPU used was an Intel Xeon Gold 6230 processor, and the system was equipped with 192 GB of RAM.

\begin{table}
    \centering
    \begin{tabular}{lc}
        \toprule
        \textbf{Parameter}             & \textbf{Value} \\
        \midrule
        Computing Infrastructure       & 40GB NVIDIA A100 GPU \\
        Optimizer                      & Adam \\
        Optimizer Params               & $\beta = (0.9, 0.999), \epsilon = 10^{-8}$ \\
        Learning rate                  & $3 \times 10^{-4}$ \\
        Learning Rate Decay            & Linear \\
        Weight Decay                   & 0 \\
        Warmup Steps                   & 200 \\
        Batch size                     & 128 \\
        Epoch                          & 5 \\
        \bottomrule
    \end{tabular}
    \caption{Hyperparameter settings for LLaMA3 8B MedReadCtrl.}
    \label{tab:hyperparameters}
\end{table}

\subsection{Human Evaluation}
\label{Sec:human_eval}

Our human evaluation was conducted by 2 medical expert evaluators (A.C. and Y.Z.).
For 3 medical datasets, We randomly sampled 10 data from the test datasets of 3 data sets, and a total of 30 data appeared in the human evaluation.
For 6 general domain datasets, We randomly sampled 6 data from the test datasets of 6 data sets, and a total of 36 data appeared in the human evaluation.
For human preference collection, we give detailed instructions to the annotators: ``\emph{You are evaluating two systems, both of which are trying to convert inputs to specific readability requirements to produce output suitable for the user. I will show you the input and output of the two systems on grade 2/5/8/11, respectively. Tell me which system's output you prefer by specifying system 1 or system 2 or tie if the quality is the same. Please explain the reason for your preference.}''.
For human rating evaluation, we asked them to follow annotation guideline in Table~\ref{tab:annotation_guideline}. 
Each time, we randomly shuffle the outputs of two systems (LLaMA3-MedReadCtrl and GPT-4), and they can choose the one that better meets the readability requirements and has higher output quality. If they think the outputs of the two systems are tied, they can choose both.
After we get judgments from two people per instance, we do not aggregate their labels before calculating the win rate but count them individually.

\begin{longtable}{p{17cm}}
\caption{Annotation Guideline for Human Evaluation} \label{tab:annotation_guideline} \\
\toprule
\textbf{Clarity and Simplicity} \\
\midrule
\textbf{Definition:} Assesses whether the explanation aligns with the reading level in terms of vocabulary, sentence structure, and complexity. \\
\textbf{Instructions:}\\
-- Grade 2: Use simple, short sentences with basic vocabulary suitable for early readers.\\
-- Grade 5: Introduce slightly complex sentence structures but remain easily understandable.\\
-- Grade 8: Include moderate complexity in language and concepts appropriate for middle school readers.\\
-- Grade 11: Demonstrate advanced vocabulary and technical accuracy suitable for high school students.\\
\textbf{Scoring:}\\
-- 1: Not suitable for the grade level.\\
-- 3: Somewhat matches the grade level but could be improved.\\
-- 5: Perfectly matches the grade level.
\\
\midrule
\textbf{Accuracy of Content} \\
\midrule
\textbf{Definition:} Evaluates if the generated text correctly explains the concept without factual errors or misleading simplifications. \\
\textbf{Instructions:} Check if the explanation maintains factual consistency while being appropriately detailed for the grade level. \\
\textbf{Scoring:}\\
--1: Contains factual errors or misrepresentation.\\
--3: Partially accurate, with minor inaccuracies.\\
--5: Fully accurate and aligned with the input content.\\

\midrule
\textbf{Consistency with Input Meaning} \\
\midrule
\textbf{Definition:} Ensures that the simplified version retains the core meaning and intent of the input text. \\
\textbf{Instructions:} Check for omissions or changes in the intended message or meaning. \\
\textbf{Scoring:}\\
--1: Major deviation from the input meaning.\\
--3: Minor deviations or slight loss of meaning.\\
--5: Fully consistent with the original input.\\

\midrule
\textbf{Overall Readability and Fluency} \\
\midrule
\textbf{Definition:} Assesses the ease of reading the text as a whole, considering grammar, punctuation, and formatting. \\
\textbf{Instructions:}\\
--Ensure no major grammatical issues or formatting inconsistencies.\\
--Grade 2 and 5: Prioritize simplicity and readability.\\
--Grade 8 and 11: Balance complexity with natural flow.\\

\textbf{Scoring:}\\
--1: Hard to read or poorly formatted.\\
--3: Readable with some flaws. Partially fluent.\\
--5: Easy to read and well-presented.\\
\midrule
\textbf{Annotation Steps} \\
1. Read the original input text and familiarize yourself with the concept.\\
2. Review the generated outputs for each reading level.\\
3. Annotate each output across the four dimensions (Clarity, Accuracy, Consistency, and Fluency \& Readability).\\
4. Assign scores for each dimension and write a brief justification (1-2 sentences) for your score.\\
5. Compare Model 1 and Model 2 based on the aggregated scores across all dimensions and provide justification.\\

\bottomrule
\end{longtable}

\subsection{AI evaluation}

To reduce the heavy human evaluation and make the evaluation easier to reproduce, we use a similar setting of our human preference evaluation for AI evaluation. 
Comparison-based feedback evaluation assesses the accuracy of LLM in deciding preferences between two responses. 
However, it is widely acknowledged that current LLMs exhibit significant \textbf{positional bias}, i.e., LLMs tend to prefer responses based on their specific position in the prompt. 
We implement a rigorous verification process to mitigate the effects of positional bias to evaluate the real capability. Specifically, given responses \(R_a\) and \(R_b\) to be compared, we obtain the comparison based on two orders, noted as \(F^c_a = F_c(R_a, R_b)\) and \(F^c_b = F_c(R_b, R_a)\). The objective scores are computed by:
\[
s = \frac{1}{N} \sum_{i=1}^N 1(L(F^c_{a,i}, F^c_{b,i}))
\]
where \(L(F^c_a, F^c_b)\) is true if and only if \(F^c_a \neq F^c_b\) and \(F^c_a, F^c_b\) align with ground-truth preference label. \(N\) is the number of test samples.
All of our experiments were conducted on the version of GPT3.5, GPT4 and Claude 3 between 25 March 2023 and 13 March 2025 by using the OpenAI’s API.10 We set temperature = 1, top\_p=1, frequency penalty = 0, and presence penalty = 0.
The prompts we used for LLM-as-a-judge (claude-3-opus-20240229 and gpt-3.5-turbo-0125) evaluation are:

\begin{table}[ht]
\setlength{\tabcolsep}{3.9pt}
\renewcommand{\arraystretch}{1.05}
\centering
\begin{scriptsize}
\begin{tabularx}{0.98\textwidth}{X} 

\toprule

You are evaluating two systems, both of which are trying to convert inputs to specific readability requirements to produce output suitable for the user.
I will show you the input and output of the two systems on grade 2/5/8/11, respectively. Tell me which system's output you prefer by specifying system 1 or system 2 or tie if the quality is the same. Please explain the reason for your preference.

**Input:**  
[\emph{input}]

**System 1 output:**  

Grade 2: [\emph{system1\_2}] 

Grade 5: [\emph{system1\_5}]  

Grade 8: [\emph{system1\_8}]  

Grade 11: [\emph{system1\_11}]

**System 2 output:**  

Grade 2: [\emph{system2\_2}] 

Grade 5: [\emph{system2\_5}] 

Grade 8: [\emph{system2\_8}]  

Grade 11: [\emph{system2\_11}]  

Please use the following JSON format for your output:  

\texttt{
\{'grade 2 preference': xxxx,  
'grade 2 preference reasons': xxxx,  
'grade 5 preference': xxxx,  
'grade 5 preference reasons': xxxx,  
'grade 8 preference': xxxx,  
'grade 8 preference reasons': xxxx,  
'grade 11 preference': xxxx,  
'grade 11 preference reasons': xxxx\}  
}

Please only output your response following the required format, and do not output any other content. Now tell me your preference and reasons:  

\\
\bottomrule

\end{tabularx}
\end{scriptsize}
\vspace{-0.05in}
\caption{Prompt of AI-evaluating readability-controlled text generation outputs.}
\label{tab:ai_evaluation}
\vspace{-0.15in}
\end{table}

\section{Data Availability}

Night datasets involved in this study are publicly available from the following links:\\
\hspace*{5mm} ReadMe: \url{https://huggingface.co/datasets/bio-nlp-umass/NoteAid-README} \\
\hspace*{5mm} MedNLI: 
\url{https://huggingface.co/datasets/bigbio/mednli} \\
\hspace*{5mm} MTSamples \url{https://github.com/babylonhealth/laymaker} \\
\hspace*{5mm} ASSET: \url{https://github.com/facebookresearch/asset} \\
\hspace*{5mm} SNLI: \url{https://huggingface.co/datasets/stanfordnlp/snli} \\
\hspace*{5mm} PAWS: \url{https://huggingface.co/datasets/google-research-datasets/paws} \\
\hspace*{5mm} WikiSmall \url{https://github.com/XingxingZhang/dress} \\
\hspace*{5mm} MultiNLI: \url{https://huggingface.co/datasets/nyu-mll/multi_nli} \\
\hspace*{5mm} MRPC \url{https://huggingface.co/datasets/nyu-mll/glue/viewer/mrpc}

\section{Code Availability}

The code is publicly available on Github: \url{https://github.com/bio-nlp/ReadCtrl}

\section{Acknowledgements}
Research reported in this study was supported by the National Center on Homelessness Among Veterans (NCHAV) and by the National Institutes of Health (NIH) under award number 1R01NR020868, and 1I01HX003711-01A1.
This study was also in part supported by NIH under award numbers R01DA056470-A1 and 1R01AG080670-01, and by the U.S. Department of Veterans Affairs (VA) Health Systems Research.
The content is solely the responsibility of the authors and does not necessarily represent NIH, VA, or the US government. 

\section{Author contributions statement}

Z.Y., H.Y. and H.T. designed the study. 
H.T., W.S. and S.S. performed the data collection. 
H.T. and Z.Y. implemented the code, and conducted
experiments. 
Z.Y. and H.T. drafted the manuscript. 
H.Y. supervised the study. 
A.C. and Y.Z. performed manual annotation and human evaluation.
All authors contributed to the research discussion,
manuscript revision, and approval of the manuscript for submission.

\section{Competing Interests}

The authors declare no competing interests.

\section*{Declaration of generative AI and AI-assisted technologies in the writing process}

During the preparation of this work, the author(s) used ChatGPT to improve the language and readability. Following the use of this tool, the author(s) carefully reviewed and edited the content as necessary and take full responsibility for the final version of the manuscript.

\bibliography{sample}

\newpage
\section*{Appendix}

\begin{longtable}{p{0.05\textwidth} p{0.92\textwidth}}
\caption{Qualitative Analysis for lower readability (e.g., Grade 2 \& 5), our model in general outperformed GPT4 in clarity, accuracy, consistency, and fluency.} 
\label{tab:qualitative_analysis_low_read} \\

\toprule

Grade2 & \textbf{Input:} X-ray visualization of the spinal cord following injection of contrast medium into the spinal arachnoid space. (README)\\
& \textbf{GPT4 output:} Taking an X-ray picture of the spinal cord after putting special dye into a space around it.\\ & 
-- Rating: \textcolor{red}{Clarity:3}, \textcolor{DeepGreen}{Accuracy:5}, \textcolor{DeepGreen}{Consistency:5}, \textcolor{red}{Fluency:4} \\&
-- Comments: \underline{I wouldn't use the term "x-ray" or even "dye" with a 2nd grader)} \\
& \textbf{Ours output:} A special picture of the spine that helps doctors see inside.\\ & 
-- Rating: \textcolor{DeepGreen}{Clarity:5}, \textcolor{DeepGreen}{Accuracy:5}, \textcolor{DeepGreen}{Consistency:5}, \textcolor{DeepGreen}{Fluency:5} \\&
-- Comments: \underline{Great. Appropriate wording.} \\

&---------------------------------------------------------------------------------------------------------------------------------------\\

& \textbf{Input:} She has cough with sputum, occasional blood streaks but no gross blood. (mednli)\\
& \textbf{GPT4 output:} She has a cough with some phlegm and a little bit of blood sometimes, but no big blood.  \\ & 
-- Rating: \textcolor{red}{Clarity:3}, \textcolor{DeepGreen}{Accuracy:5}, \textcolor{DeepGreen}{Consistency:5}, \textcolor{red}{Fluency:4} \\&
-- Comments: \uline{Phlegm may not be known to a 2nd grader; I like the effort to use terms like "little bit of" and "big" to quantify, but would have been more fluent to say "lot of" instead of "big"} \\
& \textbf{Ours output:} She has a cough and sometimes brings up yucky stuff from her lungs. Sometimes it might have a little bit of red in it, but it's not too much.  \\ & 
-- Rating: \textcolor{red}{Clarity:4}, \textcolor{DeepGreen}{Accuracy:5}, \textcolor{DeepGreen}{Consistency:5}, \textcolor{DeepGreen}{Fluency:5} \\&
-- Comments: \underline{Good.} \\

\midrule

Grade5 & \textbf{Input:} She does, however, have some tenderness in the groin bilaterally . (mtsamples)\\
& \textbf{GPT4 output:} She does have some soreness in her groin on both sides. \\ & 
-- Rating: \textcolor{red}{Clarity:3}, \textcolor{DeepGreen}{Accuracy:5}, \textcolor{DeepGreen}{Consistency:5}, \textcolor{DeepGreen}{Fluency:5} \\&
-- Comments: \underline{Not sure a 5th grader knows "groin" either.} \\
& \textbf{Ours output:} She does have some pain in the area where her legs and hips meet, on both sides. \\ & 
-- Rating: \textcolor{DeepGreen}{Clarity:5}, \textcolor{DeepGreen}{Accuracy:5}, \textcolor{DeepGreen}{Consistency:5}, \textcolor{DeepGreen}{Fluency:5} \\&
-- Comments: \underline{I really like this; it doesn't use an anatomical term but describes the location. Perfect!} \\

&---------------------------------------------------------------------------------------------------------------------------------------\\

& \textbf{Input:} Chronic inflammatory and autoimmune disease in which the salivary and lacrimal glands undergo progressive destruction by lymphocytes and plasma cells resulting in decreased production of saliva and tears. The primary form, often called sicca syndrome, involves both KERATOCONJUNCTIVITIS SICCA and XEROSTOMIA. The secondary form includes, in addition, the presence of a connective tissue disease, usually rheumatoid arthritis. (README)\\
& \textbf{GPT4 output:} This is a long-lasting illness that causes swelling and problems with the body's defense system. It mostly affects the glands that make saliva and tears, which makes it harder to produce them. The main type, known as sicca syndrome, includes two issues: dry eyes and a dry mouth. The second type happens along with another illness, often rheumatoid arthritis.  \\ & 
-- Rating: \textcolor{red}{Clarity:3}, \textcolor{red}{Accuracy:3}, \textcolor{red}{Consistency:3}, \textcolor{red}{Fluency:3} \\&
-- Comments: \uline{Worse than 2nd grade version with factual deviation; "swelling" is not really appropriate.  Also, mentioning "rheumatoid arthritis" not grade appropriate.} \\
& \textbf{Ours output:} This is a long-term illness where the body's immune system mistakenly attacks healthy glands, making it hard for people to make saliva and tears. There are two main types of this condition. One type mainly affects the eyes and mouth. The other type is linked to another condition called rheumatoid arthritis. \\ & 
-- Rating: \textcolor{red}{Clarity:4}, \textcolor{DeepGreen}{Accuracy:5}, \textcolor{DeepGreen}{Consistency:5}, \textcolor{DeepGreen}{Fluency:5} \\&
-- Comments: \underline{Good; would prefer term other than "rheumatoid arthritis."} \\

\bottomrule

\end{longtable}

\begin{longtable}{p{0.01\textwidth} p{0.92\textwidth}}
\caption{Qualitative Analysis for higher readability (e.g., Grade 8 \& 11). GPT4’s performance matches ours. However, our model tends to generate more diversified yet accurate output, while GPT4’s output is less diversified, demonstrating the effect of integrated generation leveraging both readability and language modeling.} 
\label{tab:qualitative_analysis_high_read} \\

\toprule

1 & \textbf{Input:} The inability in the male to have a PENILE ERECTION due to psychological or organ dysfunction. (README)\\
& \textbf{GPT4 output for grade 11:} The inability of a man to get an erection because of mental issues or problems with his body.
\\
& \textbf{Ours output for grade 11:} A condition in men where they are unable to get or maintain an erection, often due to a physical or mental health issue. \\ \\

&
\textbf{Input:} depression treatment drug (README)\\
&
\textbf{GPT4 output for grade 8:} a medicine for treating depression \\
&

\textbf{Ours output for grade 8:} A medicine that helps people feel better when they are sad. \\ \\ &

\textbf{Justification:} GPT4 output was essentially just rephrasing the input (replacing the input with synonyms). \\

\midrule
2 & \textbf{Input:} I will see in followup in three months . (mtsamples)\\ \\
& \textbf{GPT4 output (comments):} \\
&
\vspace{-3mm}
\begin{itemize}[leftmargin=.2in,topsep=2pt]
\setlength\itemsep{0.01em}
  \item Grade 2: I will check back in three months. \uline{-- Comments: See in followup is not the same as "check back" (which could imply a phone call, portal message, etc).}
  \item Grade 5: I will check back in three months. \underline{-- Comments: the same}
  \item Grade 8: I will check back in three months. \underline{-- Comments: the same}
  \item Grade 11: I will check back in three months. \underline{-- Comments: the same}
\end{itemize}
\\
& \textbf{Ours output (comments):} \\
& 
\vspace{-3mm}
\begin{itemize}[leftmargin=.2in,topsep=2pt]
\setlength\itemsep{0.01em}
  \item Grade 2: I will see you again in three months. \underline{-- Comments: Good.}
  \item Grade 5: I will see you again in three months to check on your progress. \underline{-- Comments: Would avoid word "progress".}
  \item Grade 8: I will see you again in three months to check on your progress. \underline{-- Comments: Good.}
  \item Grade 11: I will schedule a follow-up appointment with you in approximately three months to assess the status of your condition and determine any necessary adjustments to your treatment plan. \uline{-- Comments: Adds additional text (presumably accurate).}
\end{itemize} \\
& \textbf{Justification:} GPT4 output was similar across all levels. \\

\bottomrule

\end{longtable}

\begin{longtable}{p{0.01\textwidth} p{0.92\textwidth}}
\caption{Error Analysis of our system for hallucinations, excessive length control, and occasional instruction misinterpretation.} 
\label{tab:error_analysis} \\

\toprule

1 & \textbf{Input:} She developed a frequent dry cough about 2 weeks ago . (mtsamples)\\

& 

\textbf{Output (comments):} She got a cough a couple of weeks ago and it makes her throat feel dry. \underline{-- Comments: Suggests throat was dry, when in fact the cough was dry.} \\ \\

& 

\textbf{Input:} He was initially reluctant to undergo surgery but has had increasing amounts of chest pain episodes. (mednli)\\

& 

\textbf{Output (comments):} He didn't want to get surgery at first, but now he gets hurt in his chest a lot. \underline{-- Comments: Content not technically accurate; chest pain is not "getting hurt in his chest."  But not bad.} \\

\midrule

2 & \textbf{Input:} Chronic inflammatory and autoimmune disease in which the salivary and lacrimal glands undergo progressive destruction by lymphocytes and plasma cells resulting in decreased production of saliva and tears. The primary form, often called sicca syndrome, involves both KERATOCONJUNCTIVITIS SICCA and XEROSTOMIA. The secondary form includes, in addition, the presence of a connective tissue disease, usually rheumatoid arthritis. (README)\\ \\
& \textbf{Output for grade11 (comments):} Sjögren's syndrome is a long-standing autoimmune disorder characterized by the gradual destruction of the salivary and lacrimal glands by immune cells, leading to diminished tear and saliva production. This condition can manifest in two primary forms: the first, known as sicca syndrome, involves the simultaneous occurrence of dry eye (keratoconjunctivitis sicca) and dry mouth (xerostomia). The second form is secondary, and in addition to the primary symptoms, it is typically accompanied by the presence of a connective tissue disease, such as rheumatoid arthritis. \uline{-- Comments: Very wordy, and vocabulary unnecessarily complicated ("diminished"). Also, awkward phrasing ("the second form is secondary") which would only confuse the reader.} \\

\midrule

3 & \textbf{Input:} Description : A sample note on Rheumatoid Arthritis (mtsamples)\\ 

& \textbf{Output:} Rheumatoid Arthritis is a chronic inflammatory disorder characterized by a systemic autoimmune response, resulting in the inflammation and destruction of joints and other tissues, and is often accompanied by symptoms such as joint pain, swelling, and stiffness. \uline{-- Comments: Model misinterprets input (maybe perceiving it as instructions to create a sample note?) and generates a definition of rheumatoid arthritis.}
\\

\bottomrule

\end{longtable}

\end{document}